\documentclass[conference,compsoc]{IEEEtran}
\IEEEoverridecommandlockouts

\ifCLASSOPTIONcompsoc
  \usepackage[nocompress]{cite}
\else
  \usepackage{cite}
\fi
\ifCLASSINFOpdf
  \usepackage[pdftex]{graphicx}
  \graphicspath{{./fig/}}
  \DeclareGraphicsExtensions{.pdf,.jpeg,.png}
\else
  \usepackage[dvips]{graphicx}
  \graphicspath{{../eps/}}
\fi
\usepackage{amsmath}
\usepackage{algorithmic}
\usepackage{array}
\usepackage{color}
\usepackage{adjustbox}
\usepackage{tabularray}

\ifCLASSOPTIONcompsoc
 \usepackage[caption=false,font=footnotesize,labelfont=sf,textfont=sf]{subfig}
\else
 \usepackage[caption=false,font=footnotesize]{subfig}
\fi
\usepackage{setspace}
\ifodd 1
\else

\fi

\usepackage{url}
\newcommand{\name}{SFake} 

\begin{document}

\title{Shaking the Fake: Detecting Deepfake Videos in Real Time via Active Probes}

\ifodd 2
\author{\IEEEauthorblockN{Submission \#*** to IEEE S\&P 2025}
\IEEEauthorblockA{\textcolor{white}{ABC Inititute of Technology}}
\IEEEauthorblockA{\textcolor{white}{abc.com}}
}
\else
\author{\IEEEauthorblockN{Zhixin Xie, Jun Luo}
        \IEEEauthorblockA{Nanyang Technological University, Singapore}
        \IEEEauthorblockA{Email: \{zhixin001, junluo\}@ntu.edu.sg}
    }
\fi	

\maketitle

\begin{abstract}
Real-time deepfake, a type of generative AI, is capable of ``creating'' non-existing contents (e.g., swapping one's face with another) in a video. It has been, very unfortunately, misused to produce deepfake videos (during web conferences, video calls, and identity authentication) for malicious purposes, including financial scams and political misinformation. Deepfake detection, as the countermeasure against deepfake, has attracted considerable attention from the academic community, yet existing works typically rely on learning \textit{passive} features that may perform poorly beyond seen datasets. In this paper, we propose \name, a new real-time deepfake detection method that innovatively exploits deepfake models' inability to adapt to physical interference. Specifically, {\name} \textit{actively} sends probes to trigger mechanical vibrations on the smartphone, resulting in \textit{the controllable} feature on the footage. Consequently, \name\ determines whether the face is swapped by deepfake based on the consistency of the facial area with the probe pattern. We implement \name, evaluate its effectiveness on a self-built dataset, and compare it with six other detection methods. The results show that \name\ outperforms other detection methods with higher detection accuracy, faster process speed, and lower memory consumption.

\end{abstract}

\IEEEpeerreviewmaketitle

\section{Introduction}
In recent years, \textit{real-time deepfake} technology has received much attention~\cite{westerlund2019emergence}; it can replace a person's facial
characteristics with those of someone else, creating convincingly altered footage that maintains the original video's fluidity in real time. Such application has significantly influences various fields, such as entertainment~\cite{usukhbayar2020deepfake} and education~\cite{mutillo2023knowledge}.
Unfortunately,
deepfake has also been extensively misused in scenarios such as video calls or web conferences, where it is employed to impersonate one of the participants, deceiving others involved in the call for nefarious purposes such as financial fraud and political misinformation. As reported by CNN, a financial employee at a global company was deceived into transferring 25 million dollars to swindlers who employed real-time deepfake techniques to impersonate the firm's chief financial officer during a video conference \cite{news1}. Similar news stories of individuals and organizations being swindled by real-time deepfake videos have become increasingly common in today's digital landscape \cite{news2,news3,news4,news5,news6}. Therefore, this paper aims to design a detection method especially for thwarting real-time deepfake during video conferencing or calls on \textit{mobile device}s.

The significant losses caused by deepfake have already alerted 
the academic community. Existing works about real-time deepfake video detection mainly focus on passively seeking the 
differences between real and fake videos induced by the deepfake algorithm  \cite{mirsky2021creation}. Although these detection methods have achieved various degrees of success, they also exhibit limited generalizability across
different 
datasets~\cite{kim2021fretal,nadimpalli2022improving,chen2022ost}. 
%
%
One major reason for the poor generalization is the reliance on
detecting \textit{passively} identified features that may not be universally applicable to all deepfake algorithms and their diversified application scenarios, which has led to three main consequences.
First of all, switching to different deepfake algorithms may lead to distinct features that invalidate the trained detection algorithm~\cite{mirsky2021creation}. Moreover, deepfake algorithms are under fast developments, forcing (passive) detection methods to keep up~\cite{zhang2022deepfake}. Last but not least, diversified application scenarios (e.g., distinct backgrounds) where a deepfake algorithm is deployed can significantly affect the effectiveness of passive detection~\cite{yu2021survey}.
%

\begin{figure}[t]
    \centering
    \includegraphics[width=3.5in]{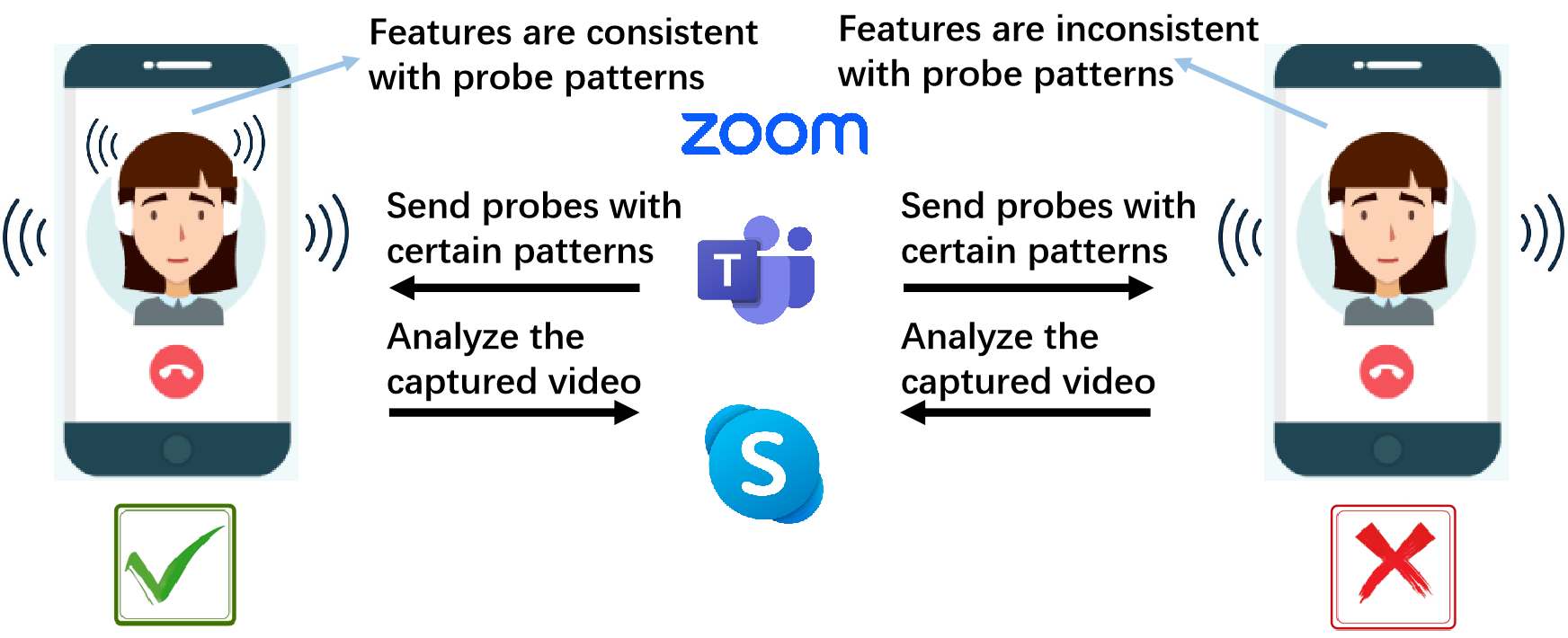}
    \caption{The overview of the {\name}. 
    The video communication software actively induces physical probes by vibrating the smartphone with certain patterns. After that, \name\ analyzes the video footage and determines the authenticity of the face by checking for the consistency between the facial area and the probe pattern.}
    
    \label{overview}
\end{figure}
The above analysis brings us to think about a research problem: \textit{can the detector actively introduce changes (or features) into videos that are i) controllable, ii) readily recognizable, and iii) affect the real and fake parts in a video distinctively?}
Our answer, also the key idea of \name, is to induce vibration to the attacker's smartphone via \textit{active probe}s. 
As shown by the basic idea of \name\ in Figure \ref{overview}, causing a smartphone to vibrate can blur the video (hence the real face in it) captured by the camera, yet such blur may not be fully adapted by deepfake. Consequently, the \textit{blur} can act as a controllable and ready-to-recognize feature. 
Although deepfake algorithms can attempt to adapt to the blur caused by vibration\footnote{According to our experience (see Section~\ref{sec_induce_blur} for details) they are indeed equipped with some level of adaptability.}, we ``turn the tables'' by forcing the attacker to keep up with the defensive side, rather than the another way around like existing passive deepfake detection methods. In particular,
we parameterize the probes to adjust the induced vibration pattern, so as to 
actively introduce controllable 
features to video footage, endowing the defensive side with 
prior knowledge 
as a key leverage.
Building upon this principle,
we propose and design \name\ as the
first deepfake detection method that may potentially offer universal defence to 
deepfake video calls on smartphones in real time.


To implement \name, we have to 
face two challenges: i) how to induce vibration to the attacker's smartphone and ii) how to detect the deepfake video 
based on the probe-induced blur in the footage. To address the first challenge, we give a detailed study on the smartphone construction and potential schemes for causing vibrations.
Our experience indicates that, though acoustic signals from external speakers can trigger vibration on smartphone~\cite{long2023side}, the 
built-in speaker of the smartphone barely causes any noticeable blur features on video footage, 
likely due to its very-low sound volume of the speakerphone.
%
Fortunately, our study also reveals
that the built-in ``vibration'' 
effects of the smartphone can be remotely activated
to introduce sufficient blur to the footage, yet at no cost of user experience.
To address the second challenge, we strategically choose not to target any specific detection algorithms; this avoids leaving vulnerabilities for potential attackers to exploit. Instead, we focus on deriving a universal \textit{blur feature sequence} that may drive virtually any detection algorithms (essentially binary classifiers), hence leaving the attackers to guess what strategy (including both vibration partten and detection algorithm) is taken by the defensive side at any point in time.
%

Overcoming these challenges, we implement the {\name}
and evaluate it on our self-built dataset with 8 brands smartphone, 15 participants and 5 existing deepfake algorithms. We compare the performance of \name\ with 6 existing deepfake detection methods. {\name} outperforms other detection methods with accuracy over 95.2$\%$, time period less than 5 seconds and memory consumption less than 450 MB. In summary, our contributions are as follows:

\begin{itemize}
\item[$\bullet$] We propose \name, the first real-time deepfake video detection scheme on smartphones that issues active probes to induce controllable blur features.
%
\item[$\bullet$] We design and implement \name\ by actively introducing physical probes, recognizing the feature patterns, and checking for their consistency.
%


\item[$\bullet$] We evaluate {\name} on our self-built dataset. The results show that {\name} outperforms other six detection methods in accuracy, process speed and memory consumption.
\end{itemize}
When put into practical use, {\name} can be integrated as a specific security functionality into video communication applications. When the process of video communication begins, the application utilizes the \name\ for deepfake detection, and alerts the user if necessary.

\begin{figure}[t]
    \centering
    \includegraphics[width=3in]{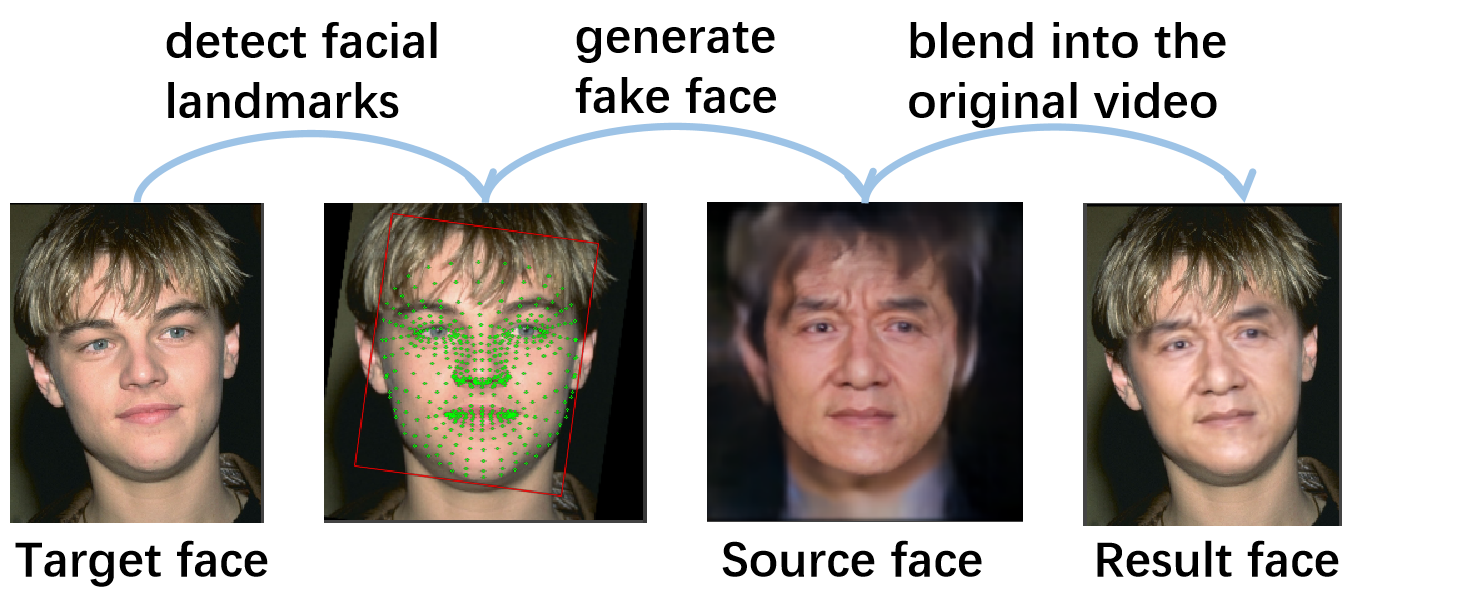}
    \caption{The main steps of FSA.}
    \label{fake_face_background}
\end{figure}
\section{Background}\label{sec_background}
In this section, we present the background knowledge of i) the threat model of our detection method, ii) the workflow of deepfake algorithms and limitations of the existing detect methods, iii) the intuition of our detection method, and iv) the related components of the smartphone.

\subsection{Threat Model}\label{sec_threat_model}
\textbf{Deepfake attack model}. The scenarios of deepfake attacks mainly involve 
video communication where the participants use digital video and audio to communicate. 
Generally speaking, deepfake attacks focuses on 
replacing the attacker's 
faces with that of a legitimate participant, aiming to deceive other participants during video communication. 

\textbf{Defense goals}. 
\name\ should detect the deepfake-generate videos in different scenarios as accurately as possible. Besides that, 
The detection process should be resource-saving as it is designed for the mobile device.

\textbf{Defender's capability}.
Like other facial recognition functionality, {\name} requires the user to remain as still as possible for a short period (e.g., 4 s) for detection. {\name} requires cameras with resolutions of 1920*1080 pixels or higher, and they must support at least 2x zoom capabilities. If a phone's front camera does not meet these requirements, the rear camera can be used as an alternative. {\name\ should to some extent be able to control the hardware, such as zooming out the camera and playing the ``vibration'' sound effect during the detection process.

\subsection{Limitation of Existing Methods}
In this part, we briefly illustrate the basic principle of the face swap algorithm (FSA) and the limitations of existing detection methods. The main steps of the FSA are summarized in Figure \ref{fake_face_background}.  
The FSA detects the face area of the target face and extracts several tens of facial landmarks. Based on the facial landmarks, the FSA generates the fake face based on the source face and blends it with the the background of the real image, thus get the result face. To generate a fake video, FSA applies the above steps to each frame of the video and merge the generated fake images into a fake video.

\begin{figure}[t]
    \centering
    \vspace{-0.2cm} 
    \begin{adjustbox}{margin=15pt 0pt 0pt 0pt}
        \includegraphics[width=3.8in]{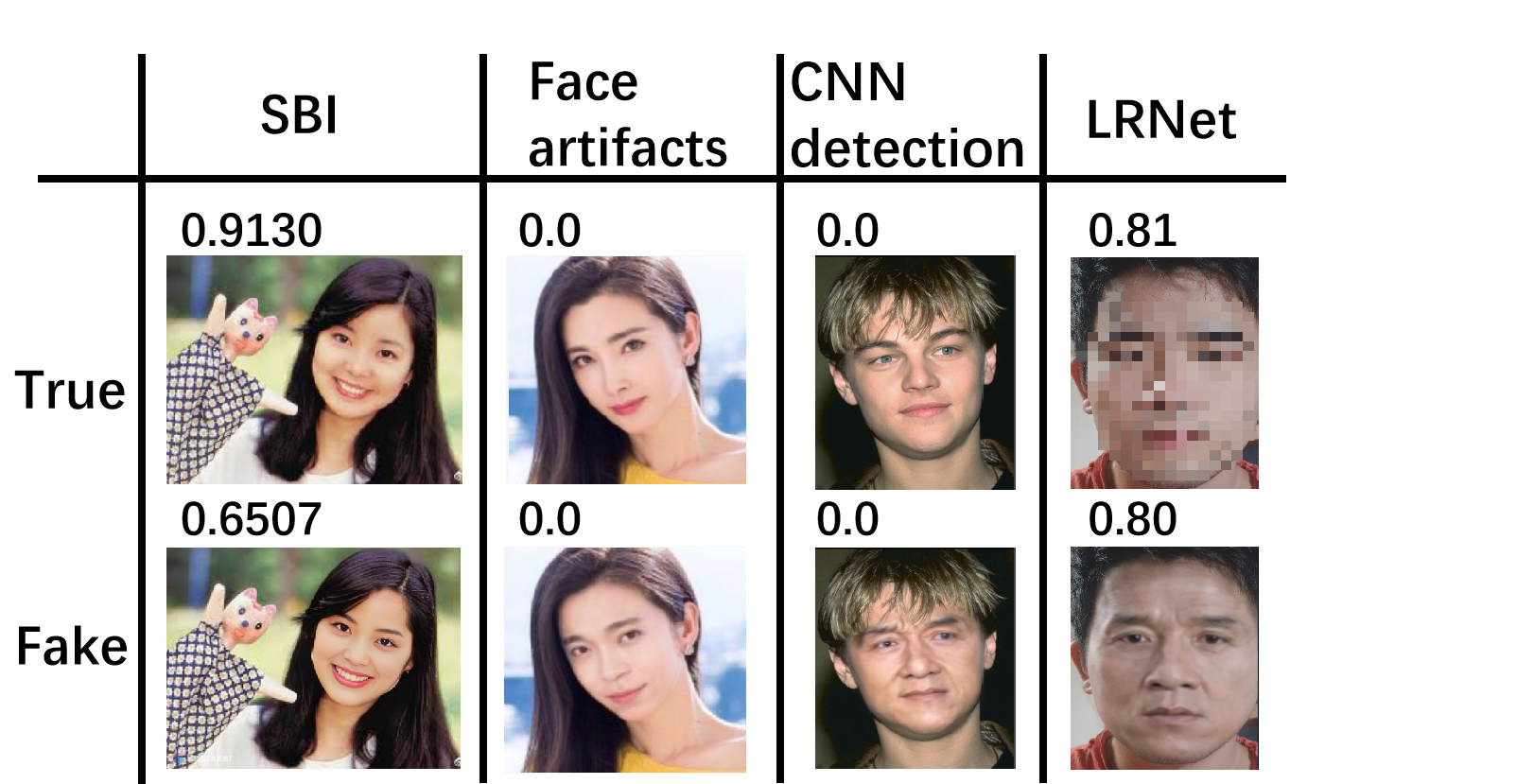}
    \end{adjustbox}
    \caption{The four cases where the detection methods fail, with real images/videos, their corresponding fake images/videos, and their fakeness scores given by the detection model.
    }
    \label{fakeness}
\end{figure}

However, due to the imperfection of the current deepfake algorithms, the forged video contains unnatural traces caused by face swap, which can serve as a feature for distinguishing between real and fake videos. The general idea of existing detection methods is to extract these features passively. The typical features include the unregular artifacts on the fake face caused by the face synthesis step\cite{li2018exposing,vairamani2022analyzing,dhanaraj2024face,li2022artifacts,chen2023watching,chhabra2023low}, the blending traces at the splices of the fake face and the head at the blending step \cite{liu2024enhancing,shiohara2022detecting,hsu2021deepfake,straub2019using,masi2020two}, the fingerprints of the corresponding generative network that is inherently contained in the fake video \cite{yu2019attributing,marra2019gans}, the temporal mismatch between the physiological signal and footage \cite{zhou2021joint,yang2019exposing,agarwal2019protecting,conotter2014physiologically,korshunov2018speaker}, and so on. After selecting the specific features that can represent the differences between real and fake videos, a network is trained based on the well-established datasets \cite{zhou2021face,he2021forgerynet,jiang2020deeperforensics,dolhansky2020deepfake,rossler2019faceforensics++} to detect the subtle differences mentioned above. However, the existing detection methods lack robustness due to the diversity of deepfake algorithms, their rapid development, and the variety of usage scenarios.
G. Pei et al. \cite{pei2024deepfake} compiles statistics from 14 papers on deepfake detection published in recent years, showcasing the results of cross-validation performed on datasets like DFDC \cite{dolhansky2020deepfake}, Celeb-DF, \cite{li2020celeb}, Celeb-DFv2 \cite{li2019celeb}, and DeeperForensics-1.0 \cite{jiang2020deeperforensics}. On the DFDC dataset, which has a larger size and more deepfake algorithms incorporated, the best accuracy among the 14 detection methods achieved 90.3$\%$, while the lowest accuracy was 67.44$\%$, which shows there is still room for improvement in the generalizability of current work.

To further illustrate the unstable performance of the existing detection methods, we show four cases where the detection systems fail to determine the authenticity of the face in Figure \ref{fakeness}. We input four pairs of images/videos into four models: SBI~\cite{shiohara2022detecting}, Face artifacts~\cite{li2019exposing}, CNN detection~\cite{wang2019cnngenerated} and LRNet~\cite{sun2021improving}. In these four cases, the models incorrectly assign a higher fakeness score to the real images/videos than to the fake ones. This might be because we use an online face swap tool~\cite{RemakerAI2023} to generate fake content not contained in the mainstream deepfake datasets and unseen to the detection model. The above examples show that the passively extracted features are uncontrollable and change with different deepfake algorithms, which are readily manipulable to attackers.

\begin{figure}[t]
    \centering
    \vspace{-0.2cm} 

    \includegraphics[width=0.3\textwidth]{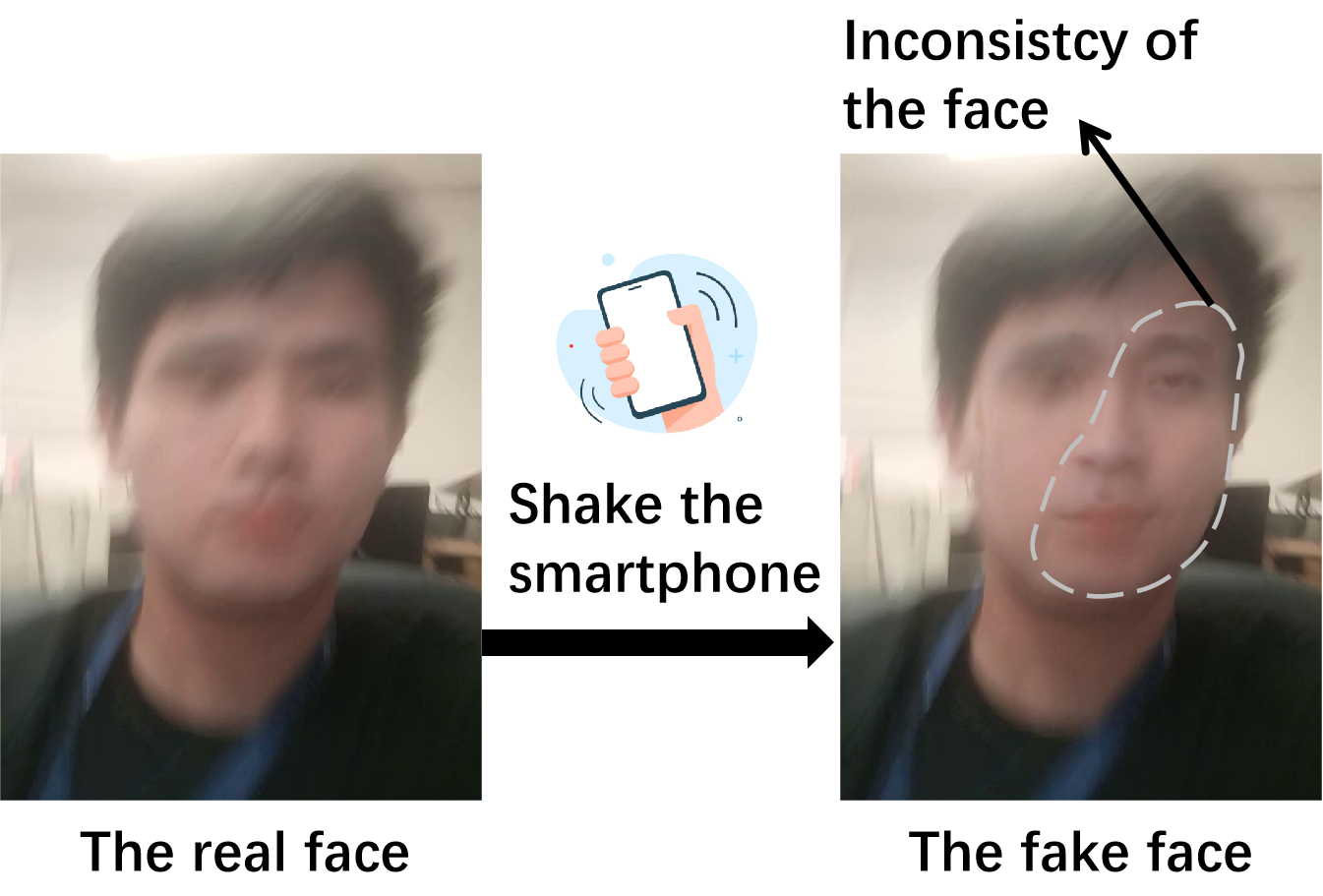}
    \caption{The intensive shake of the smartphone causes heavy distortion on the target face, which in turn causes inconsistency on the result face (the circled part in the right picture).
    }
    \label{intuitive}
\end{figure}

\subsection{Intuition of Our Method}
As passively extracting features depends much on the attacker's strategies and has limited robustness, we think about how to actively introduce the features that are controllable to the defender and insensitive to alteration of the attacker's strategies. The very straightforward idea is to shake the attacker's smartphone intensively, which blurs the captured video. As the source face remains unchanged, the different sharpness of the source face and target face is likely to cause inconsistency on the result face, no matter what algorithm the attacker uses. As shown in Figure \ref{intuitive}, the circled part of the fake picture on the right is sharper than other parts, which can even be observed by raw eyes, not to mention the well-designed detection algorithm. Therefore, the shake-caused blur can act as a feature to detect the fake faces. However, the defender cannot shake the attacker's smartphone as they are typically not in the same space. Therefore, we conduct a comprehensive study of the smartphone's structure to find methods that can make smartphones move controllably even far away from it. Finally, we find the ``vibration'' sound effect meets our requirement, because it not only generates mechanical vibrations sufficiently large but also is remotely controllable.

\subsection{Related Smartphone Components}

\textbf{``Vibration'' sound effect}. 
``Vibration'', a widely used sound effect of the smartphone,
is triggered by a built-in motor. Traditionally, 
with an off-centered weight attached, 
the center of gravity moves as the motor spins.
Recently, linear motors which utilizes the electromagnetic effect as drive source are 
also commonly used in smartphone applications. 
Regardless of their running principles, the motor inside the smartphone can be controlled by the software, so the period and duty cycle of the vibration can be customized, enabling different vibration patterns for various applications. 

\textbf{Refocusing of the camera}. In order to keep the captured image clear, the camera needs to adjust the position of the optical system in real time, so that the imaging position of the object falls on the sensor pixel array. This process is called focusing. When the relative position between the camera and the object changes, the camera needs to refocus to make the object clear. The image will become blurred until the refocusing is completed, which is  potential to serve as the criteria for determining whether the smartphone has moved. 

\begin{figure}[t]
    \centering
    \subfloat[][]{\includegraphics[width=0.2\textwidth]{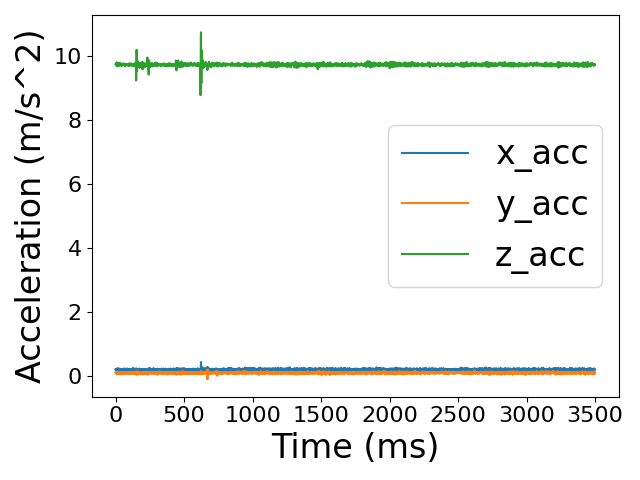}}
    \subfloat[][]{\includegraphics[width=0.2\textwidth]{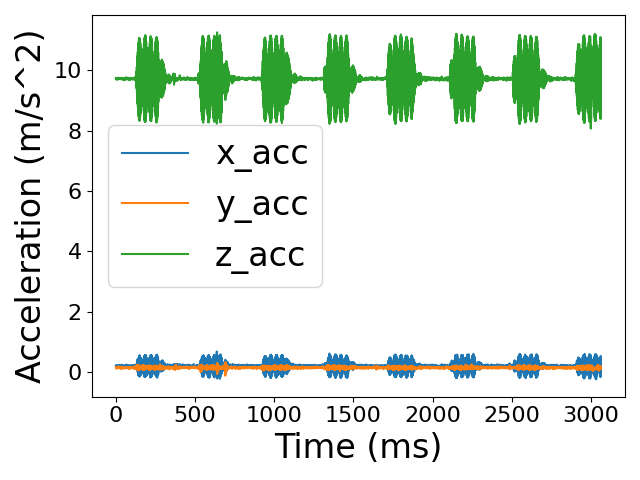}}\\	
    \subfloat[][]{\includegraphics[width=0.32\textwidth]{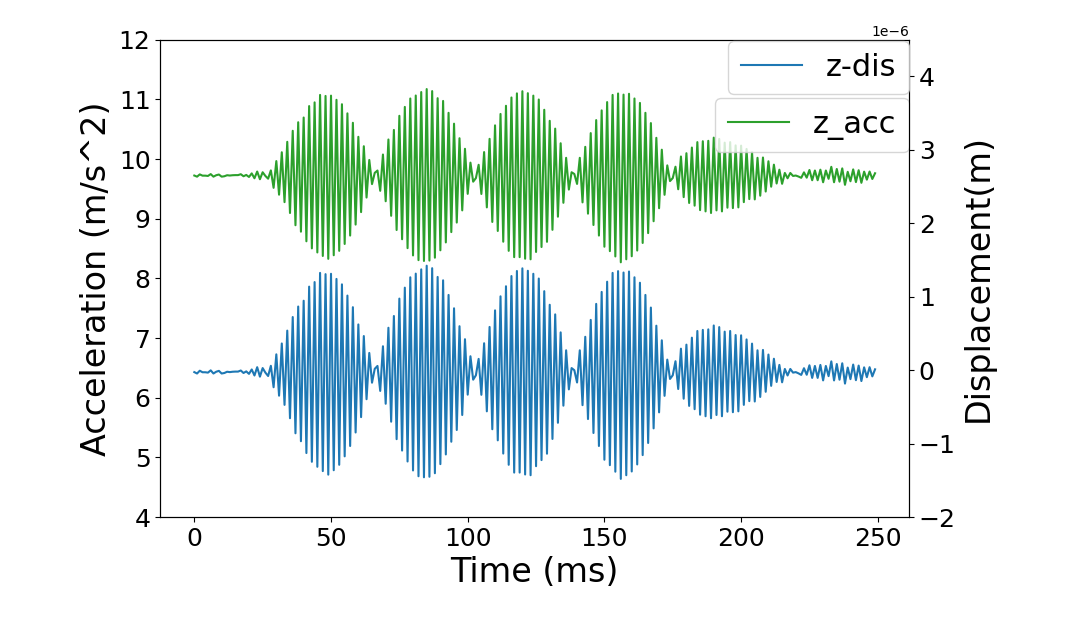}}		
    \caption{(a) The three-axis acceleration when playing a song at maximum volume. (b) The three-axis acceleration when playing the ``vibration'' sound effect. (c) The z-axis acceleration and the corresponding calculated z-axis displacement.}
    \label{acc}
\end{figure} 

\section{Feasiblity Analysis}\label{sec_fea_ana}
Before we present the design of \name, in this section, we illustrate the feasibility of inducing slight movements to the smartphone as the controllable physical probe and blurring the video footage as the readily recgnised feature by playing the ``vibration'' sound effect.
\subsection{Induce Movement to Smartphone}

To remotely vibrate the smartphone, we are initially inspired by the idea in SideEye attack\footnote{In the SideEye attack, the attacker exploits the smartphone's camera to infer the sound played by a speaker located on the same surface as the device.}\cite{long2023side} of using the smartphone's speaker to generate a sound wave that shakes the camera. However, based on our experiments, the phone's volume is too low to cause sufficient vibration. We collect the acceleration data of the smartphone by its inertial measurement unit (IMU) while it is playing a piece of music at its maximum volume, and the result in Figure \ref{acc}(a) shows that the sound wave hardly introduces movement to the smartphone. Therefore, we conduct a comprehensive study and attempt to identify methods that can substantially vibrate the smartphone. We finally turn to the ``vibration'' sound effect. Unlike other sound effects where the smartphone's movement is merely a byproduct, the ``vibration'' sound effect is generated by a motor specially designed for mechanical movement. As a result, the ``vibration'' sound effect causes a much greater amplitude of movement, as shown in Figure \ref{acc}(b). To more quantitatively describe the motion of the smartphone, we calculate the displacement of the smartphone in the z-axis (as the movement primarily occurs along the z-axis) according to Eq. \ref{acc_dis}
\begin{align}
\begin{cases}
    V_{i+1} = V_i + a_i*\Delta t \\
    D_{i+1} = D_i + V_i*\Delta t \\
    D_0 = V_0 = 0
\end{cases}
\label{acc_dis}
\end{align}

where $V_i$ and $D_i$ represent the speed and displacement of the smartphone along the z-axis at the $i^{th}$ moment respectively. Figure \ref{acc}(c) shows the results. The displacement curve has a similar trend to the acceleration curve and reaches its maximum of about 1.5 $\mu$m.

In addition,
vibration-induced movements are easier to control, thus enriching the feature patterns and increasing the randomness of detection. For example, the detector can randomize the period and duty cycle of vibration before each detection. As a result, even if an attacker knows the detection method in advance, it would be challenging to determine the specific detection parameters, making it difficult to bypass the defense.

\subsection{Induce Blur to Video}\label{sec_induce_blur}
\textbf{The structure of the camera and the blurriness under ideal conditions.} In this part, we briefly introduce the fundamental structure of the imaging system and illustrate the theoretically minimum blurriness. An imaging system is simplified to consist of an object, a lens, and a photosensitive sensor array, as shown in Figure \ref{camera}. All light rays originating from one point of the object passing through the lens converge to the image point. To get a sharp image, the sensor array should adjust its position along the optical axis until it contains the image point. Under ideal conditions, one object point maps precisely to one image point, shown as the black point on the right side of Figure \ref{camera}. However, when the object point vibrates, the image point also vibrates, making light rays not focus ideally on the sensor array. The imaging result is no longer an ideal geometric point but becomes a circle with a physical size in space, termed the ``circle of confusion'' (CoC), shown as the blue and green circles in Figure \ref{camera}. The size of the CoC represents the degree of the image's blurriness.

Even with the smartphone being stationary and focused, an ideal camera exhibits inherent blurriness due to several factors. First, the sensor array's pixel size limits the imaging resolution. Even if the light rays focus ideally on the sensor array, the image is not a perfect point but a rectangle with the size of a pixel. Second, due to the diffraction of light, an object point is naturally imaged to a circle known as the Airy disk \cite{openstax2016collegephysics}, whose size theoretically limits the upper bound of imaging resolution with a particular aperture \cite{williams2009transmission}. We take the camera on Xiaomi Redmi 10X \cite{GSMArena2020XiaomiRedmi10X} as an example to calculate the ideal resolution. First, the size of each pixel is 0.8 $\mu$m, as referred to in the specifications. Second, according to Rayleigh criterion\cite{williams2009transmission}, the diameter of the Airy disk is 0.87 $\mu$m with the aperture of f/1.8. Therefore, the smallest CoC diameter of the camera on Xiaomi Redmi 10X is limited by its sensor arrays, which is 0.8 $\mu$m. 


\begin{figure}[t]
    \centering
    \includegraphics[width=3.5in]{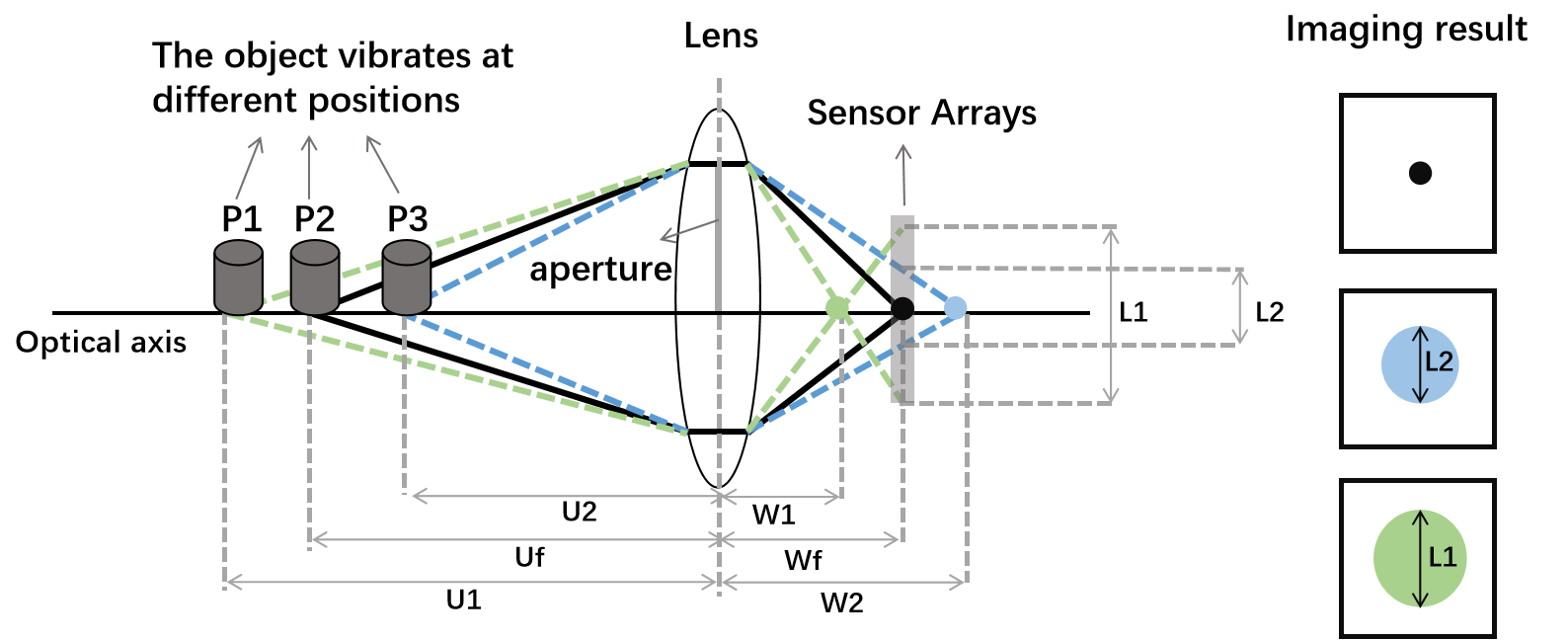}
    \caption{Illustration of three imaging situations: (1) the sensor array contains the image point, and the object point is precisely imaged to the image point; (2) the sensor array is so close to the lens that the object point is imaged to a circle of confusion (shown as the blue light rays and circle); (3) the sensor array is so far to the lens that object point is also imaged to a circle of confusion (shown as the green light rays and circle).}
    \label{camera}
\end{figure}

\textbf{The theoretical analysis of blurriness induced by vibration.} As illustrated in Section \ref{sec_background}, the movement along the z-axis leads the camera to refocus and introduce blur on the video. In this part, We further explore the quantitative relationship between the movement of the smartphone and the degree of blurriness represented by the CoC's radius. We assume the object distance as $u$, the image distance as $w$, the focus length as $f$, and the aperture as $F$ (then the size of the aperture equals $\frac{f}{F}$ by definition) and the diameter of the CoC as $r$. According to the thin lens equation
\begin{equation}
    \frac{1}{f} = \frac{1}{w} + \frac{1}{u} \Rightarrow w = \frac{f*u}{u-f},u=\frac{f*w}{w-f}
    \label{thin_law}
\end{equation}
We first consider the situation in which the object point moves towards the lens, represented as the blue lines and blue circle in Figure \ref{camera}. According to the geometric properties of similar triangles  

\begin{equation}
    \frac{w_2-w_f}{L_2} = \frac{w_2}{F} \Rightarrow w_2 = \frac{f*w_f}{f-F*L_2}
    \label{geo_eq}
\end{equation}

We substitute Eq. \ref{thin_law} to Eq. \ref{geo_eq}, and obtain the equation set

\begin{align}
\begin{cases}
    w_f = \frac{f*u_f}{u_f-f} \\[3pt]
    u_2=\frac{f*w_2}{w_2-f} \\[3pt]
    w_2 = \frac{f*w_f}{f-F*L_2}
\end{cases}
\label{eq_set}
\end{align}

Solving the equation set, we get the expression of $L_2$, the radius of CoC caused by the object point moving towards the lens
\begin{equation}
    L_2 = \frac{f^{2}*(u-u_2)}{F*(u-f)*u_2}
    \label{L2}
\end{equation}
By analogy, when the object point moves away from the lens, the radius of CoC can be represented by the expression 
\begin{equation}
    L_1 = \frac{f^{2}*(u_1-u)}{F*(u-f)*u_1}
    \label{L1}
\end{equation}

To illustrate the impact of vibration on the degree of image blurriness, we take the Xiaomi Redmi 10X smartphone as an example where $f$ is 26 mm, and $F$ is $\frac{f}{1.8}$. According to Eq. \ref{acc_dis}, a typic value of the maximum vibration amplitude is 1.5 $\mu$m. We assume the distance $u$ between the lens and the object is 10 cm, and then $u_1$ and $u_2$ are respectively $10 cm + 1.5 \mu m$ and $10 cm - 1.5 \mu m$. According to Eq. \ref{L1} and Eq. \ref{L2}, $L1$ and $L2$ are approximately equal to 0.095 $\mu$m, which represents the blurriness caused by vibration. Compared to the inherent error of 0.8um in an ideal camera, the error caused by vibration is about one-tenth of it. That means the light rays initially contained within a single pixel’s sensor now partially affect the surrounding pixel, leading to mutual interference between the pixels. Since the edges of color blocks experience the most significant color changes, the blurriness caused by vibration is primarily manifested in the slowing of color transitions at the edges and the widening of the edges. When the camera vibrates rapidly along with the smartphone, the camera does not have enough time to complete the focusing process, resulting in a continuous defocus state, leading to sustained blurriness in the video.

\begin{figure}[t]
    \centering
    \includegraphics[width=0.4\textwidth]{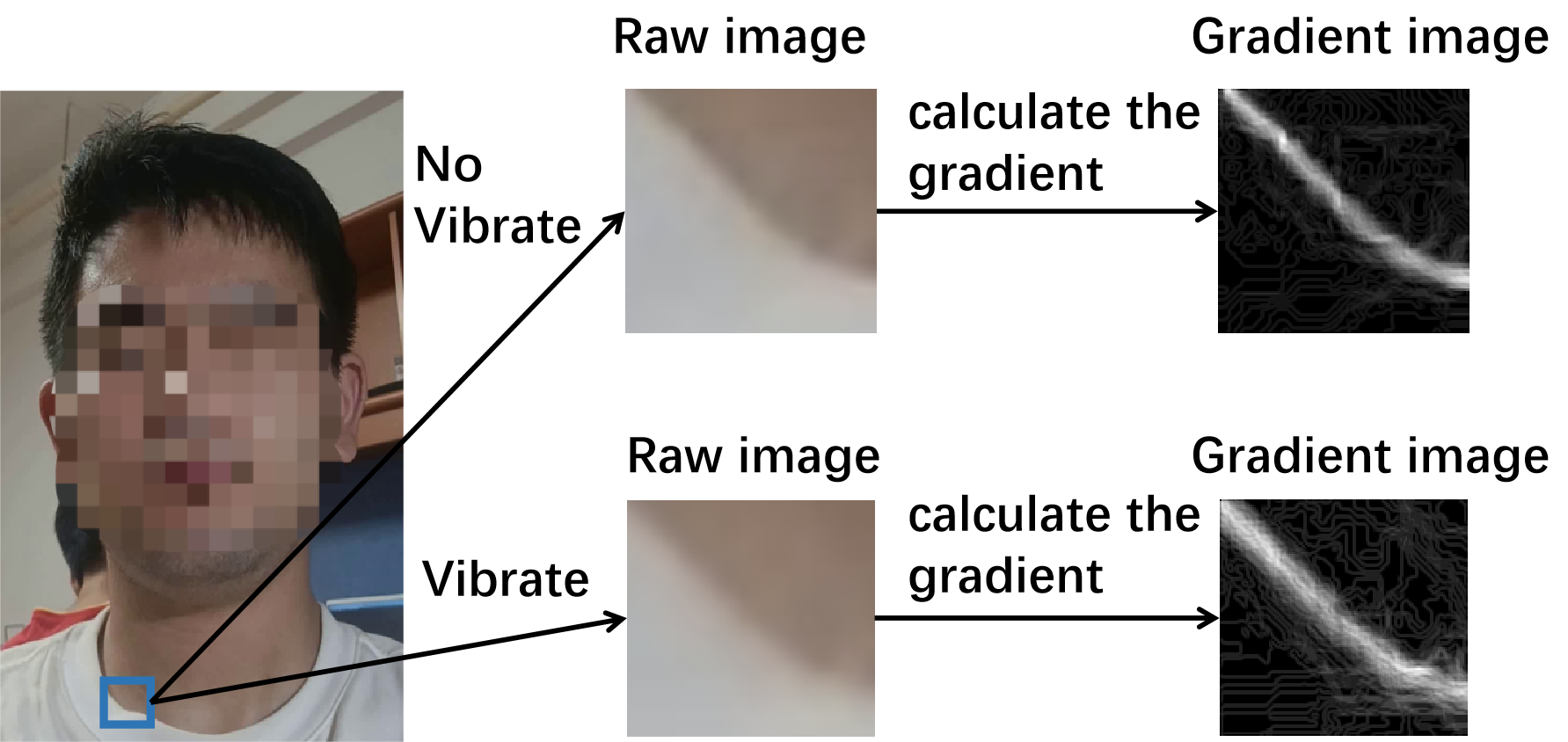}
    \caption{We compare the collar area of the video with and without the vibration.}
    \label{fesibility_blur}
\end{figure}

\textbf{The experimental verification of blurriness induced by vibration.} To validate the above theoretical hypothesis, We record an 8-second video with the vibration period of 2 seconds and the duty cycle of 0.3. We observe the vibration of the subject's collar in the video as an example. The result is shown in Figure \ref{fesibility_blur}. When the smartphone does not vibrate, the collar area is sharp and the edge between the skin and the clothing is distinct. When the smartphone vibrates, the skin and clothing are blended along the edge of the two parts. We calculate the gradient of two images, which can better represent the distribution of edges. The result shows that gradients of the vibrating image are more widely distributed in space, whereas they are more concentrated in the non-vibrating images.

Therefore, when the camera defocuses, the edges of the imaging will be less sharp, resulting in a more gradual variation in pixel values. We can use variance to measure the degree of variation of an image, which is calculated by Eq. \ref{variance}
\begin{equation}
\sigma^2 = \frac{1}{N} \sum_{i=1}^{N} (x_i - \mu)^2 \label{variance}
\end{equation}
Here, $\sigma^2$ represents the variance, N is the number of pixels, $x_i$ represents the value of each pixel and $\mu$ is the mean of pixel values. We calculate the variance of the gradient of the collar area in Figure \ref{fesibility_blur} frame by frame in the 8-second video and collect acceleration data from the IMU during video recording. The relationship between IMU readings and video variance is shown in Figure~\ref{vib_var}. When the smartphone vibrates, the variance of the images decreases, which means the footage gets blurred. The experiment shows that vibration blurs the video, which can be detected by variance sequence.

\begin{figure}[t]
    \centering
    \vspace{-0.5cm} 
    \includegraphics[width=0.3\textwidth]{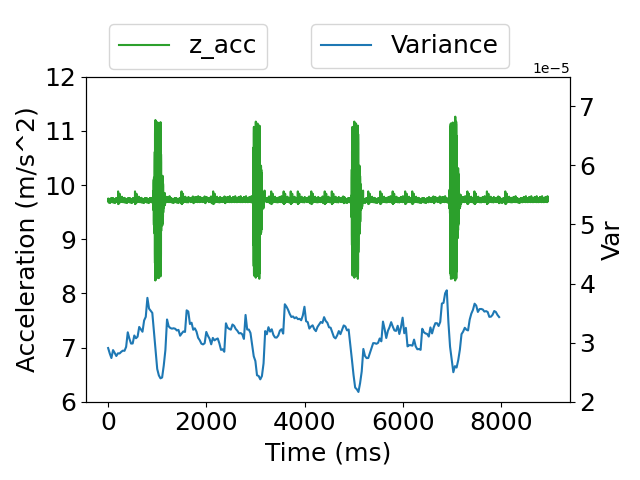}
    \caption{The relationship between the video variance sequence and the vibration.}
    \label{vib_var}
\end{figure}

\textbf{How deepfake algorithm deals with blurriness.} FSA exhibits a certain adaptability for blurry images. For example, when it performs a face swap on a blurry target image, the result image is also quite blurry, as shown in Figure \ref{blur_process}(a). However, the vibration caused blurriness is too small to be recognized by FSA. To prove this conclusion, we select a facial area of the first frame of the 8-second video and use Gaussian filtering to blur it, simulating the effect of vibration on the video. To choose the proper size of the Gaussian kernel, we compare the variance of the video with and without vibration. we find that when the kernel size is between 1 and 3, the blurriness effect of Gaussian filtering is similar to the vibration. Therefore, we Gaussian filter the same sample image 100 times with random size of kernel and apply the FSA algorithm to generate 100 fake faces. The variance of real and fake faces are respectively shown as blue and green curves in Figure \ref{blur_process}(b). The trends of the green and blue curves are not consistent, which indicates that FSA cannot recognize the slight variation of blurriness caused by vibration.

The above analysis leads to a conclusion: for the real face, artificially controlled vibrations cause a regular blurriness which can be reflected by the variance. However, the FSA can not precisely determine the vibration-caused blurriness of the target image and the fake face does not exhibit regular blurriness. Therefore, vibration-caused blurriness can serve as the feature that differentiates the real and fake videos.

\begin{figure}[t]
    \centering
    \vspace{-1cm} 
    \subfloat[][]{\includegraphics[width=0.20\textwidth]{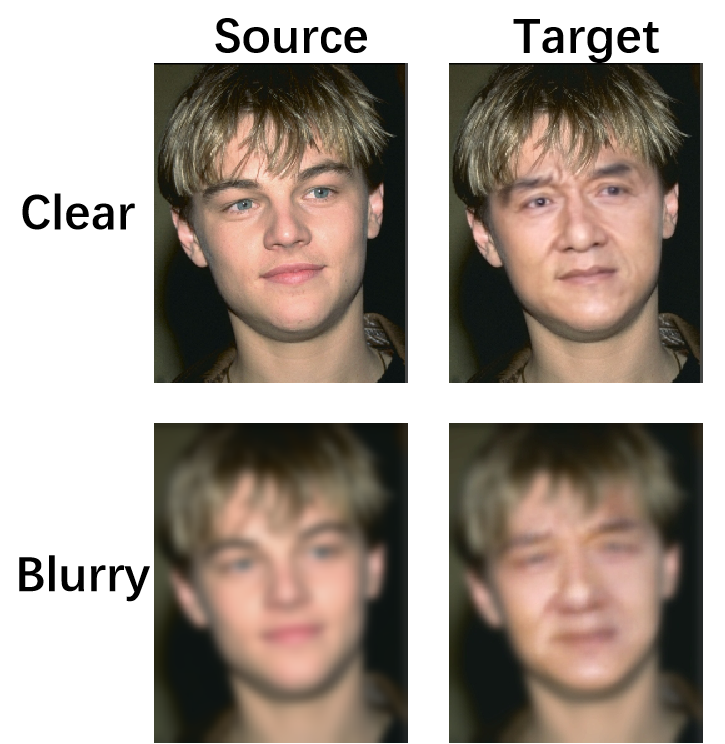}}
    \subfloat[][]{\includegraphics[width=0.3\textwidth]{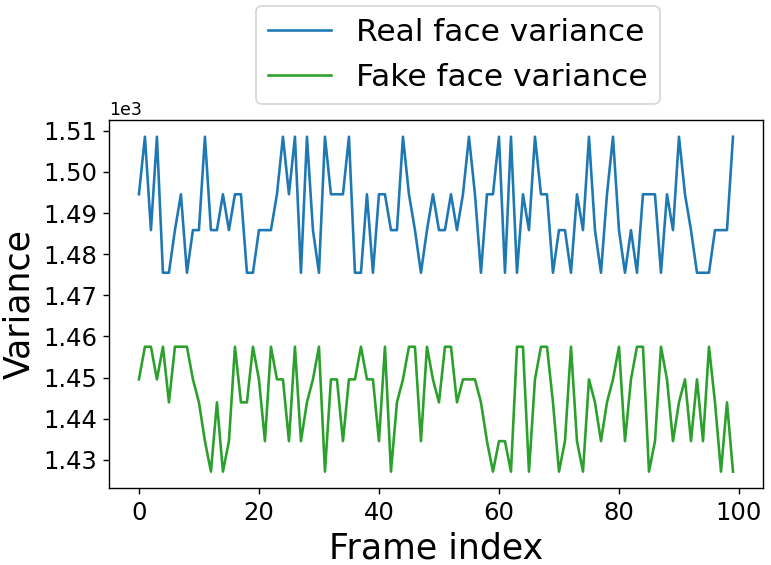}}\\		
    \caption{(a) The comparison between clear and blurry images demonstrates that FSA has a certain adaptivity to process blurry images. (b) The variance of clear and blurry image sequences when the Gaussian kernel size is randomly distributed between 0 and 3. The blurriness caused by vibration is too slight for FSA to recognize and process.}
    \label{blur_process}
\end{figure}

\begin{figure*}[t]
    \centering
    \includegraphics[scale=0.3]{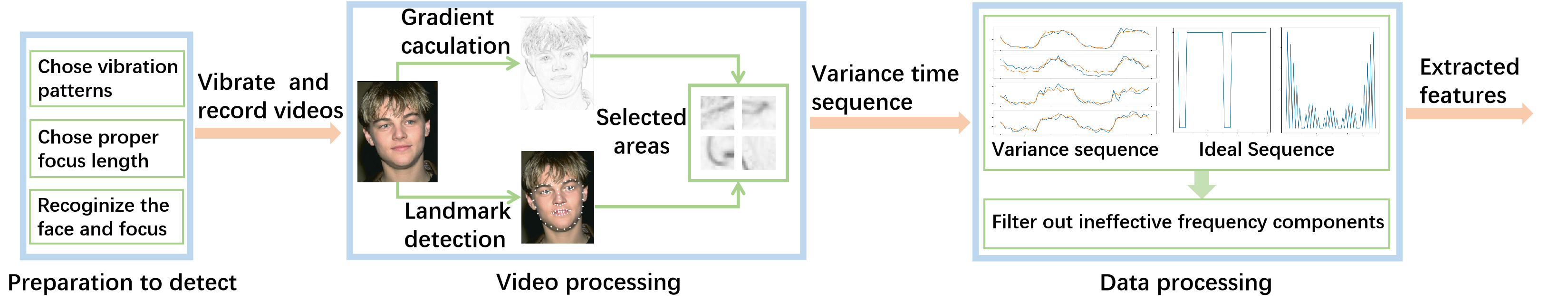}
    \caption{The workflow of the {\name}.}
    \label{workflow}
\end{figure*}

\begin{figure}[t]
    \centering
    
    \subfloat[][]{\includegraphics[width=0.22\textwidth]{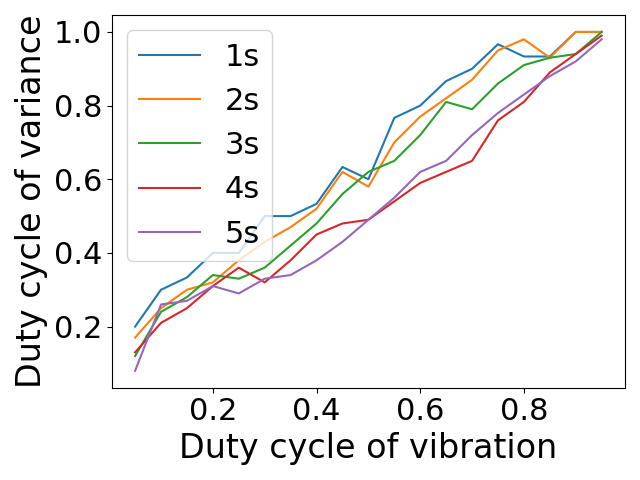}}
    \subfloat[][]{\includegraphics[width=0.22\textwidth]{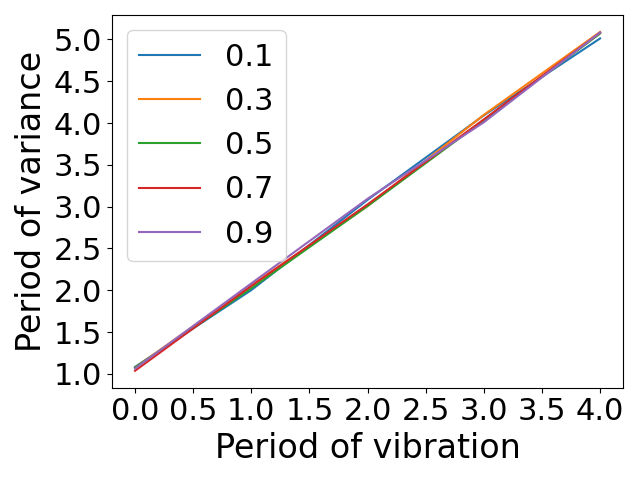}}\\
    \caption{(a) The relationship between the duty cycle of variance and vibration across different periods. (b) The relationship between the period of variance and vibration across different duty cycles.}
    \label{vib_blur}
\end{figure}

\section{System Design}\label{sec_sys_des}


We design the {\name} with three main steps, as shown in Figure \ref{workflow}. First, the {\name} sets the artificially designed or randomly generated vibration pattern. The {\name} also sets a proper focus length of the camera and focuses on the facial area, as it can influence the degree of the blurriness of the video. Meanwhile, the smartphone starts to vibrate and capture the facial area. Afterward, {\name} selects areas most likely to reflect the vibration pattern by detecting the facial landmarks and analyzing the image's gradient information. Finally, we obtain the feature by calculating the variance of these areas frame by frame and filtering out the noise. We determine the authenticity of a video by assessing whether its features reflect the vibration pattern. 

\subsection{Preparation to Detection}
\textbf{Selection of vibration patterns.} \name\ can set the period and duty cycle of the vibration to configure its pattern. Theoretically, the variance sequence should be directly proportional to the vibration pattern. However, the sensitivity of the variance sequence to vibration is limited. For example, if the duty cycle of vibration changes from 0.50 to 0.51, the variance sequence may not change so much. To quantitatively explore the relationship between the vibration patterns and variance sequence, we set the vibration period from 1 to 5 seconds with a step size 1 second and the duty cycle from 0 to 1 with a step size 0.05. For each vibration pattern, we record a video, calculate the variance sequence frame by frame, and utilize the middle value of the variance sequence as a threshold to calculate its duty cycle. The experiment result is shown in Figure \ref{vib_blur}. In general, the variance sequence can reflect the pattern of vibration. As for period, Figure \ref{vib_blur}(b) shows that under different duty cycles, the period of variance sequence equals that of vibration. As for the duty cycle, the overall trends of variance sequence reflect the duty cycle of the vibration, but the details are not entirely accurate. For example, when the period of vibration is 1s or 2s, the duty cycle of variance decreases with that of vibration increasing from 0.45 to 0.55, as shown in Figure \ref{vib_blur}(a). Therefore, we coarsely set three options for the duty cycle: 0.2, 0.5, and 0.8. The blurriness of the videos corresponding to these three settings exhibits significant differences, ensuring that the variance sequence can reflect the vibration patterns. Additionally, since the variance sequence almost aligns with the vibration periods, we do not set specific options for the period to increase the diversity of vibration patterns.

From an implementation perspective, the existing mobile framework provides feasibility for controllable vibration. Manual vibrations can be triggered using the Vibrator class's \textit{VibrationEffect} for more nuanced patterns. 

\textbf{Camera related preparation}
As analyzed in Section \ref{sec_induce_blur}, inducing blur through the smartphone's vibration relies on the camera being in focus. Therefore, before initiating the vibration process, we need to make the camera focus on the face. We use a face recognition algorithm provided by Dlib~\cite{dlib}, a widely used machine learning toolkit, to detect the facial area in the captured video. After that, we operate the camera and focus on the face. According to Eq. \ref{L1} and Eq. \ref{L2}, the blurriness caused by vibration is related to the camera's focal length. As the focal length increases, the radius of the CoC also increases, resulting in more severe blur. However, as the focal length increases, the field of view (FOV) decreases, which means the camera may not be able to capture all areas of the face. Therefore, we need to adjust the focus length to make it as large as possible while still being able to capture the entire facial area. From experience, the focus length can be set around 50mm, effectively doubling the images, which can induce sufficient vibration while capturing the entire facial area.

From an implement perspective, in Android environment, the focus distance can be adjusted by setting \textit{LensFocusDistance} in the \textit{CaptureRequest.Builder}. After that, the application can choose the focus area by setting the value of \textit{MeteringRectangle} and applying it to the class \textit{CaptureRequest.Builder}. 

\subsection{Video Processing}\label{sec_video_process}
The fundamental idea of \name\ is to determine the authenticity of a video by comparing the changes in video blurriness and vibration patterns. To accurately measure the blurriness, we calculate the gradient of the image. However, to conserve computational resources, we only select several representive areas to perform gradient calculation.

\textbf{Gradient calculation}. As previously analyzed, the blurriness caused by vibration primarily manifests at the edges of color blocks in the footage. In previous discussions, we directly use the variance of the image as a measure to assess the level of blurriness for further analysis. This method is applicable under good shooting conditions because the distribution of pixel values inside the color blocks is relatively uniform with slight variation in such cases. However, when the shooting conditions are poor, noise may occur within the color blocks due to lighting fluctuations, sensor noise, etc. The noise is widespread throughout the entire image and constantly changes over time, thereby affecting the value of the variance, making the variance not solely affected by the vibration.

To eliminate the influence of noise, we apply gradient processing to the image and remove areas with minimal gradient values to prevent the noise within the color blocks from interfering the variance calculation. we calculate the gradient of the image by Eq. \ref{gradient}
\begin{equation}
g[i,j] = \frac{1}{9} \sum_{u=i-1}^{i+1}\sum_{v=j-1}^{j+1} abs(f[u,v]-f[i,j]) \label{gradient}
\end{equation}
where $f[i,j]$ refers to the pixel value at $i^th$ row and $j^th$ column, while $g[i,j]$ denotes the grayscale value at the same coordinate. After that, we zero out the gradients smaller than one-tenth of the maximum gradient value. Figure \ref{three_pic}(a) shows the raw and gradient image. To demonstrate the effect of the gradient calculation, we record a 4-second video with the vibration pattern of 1 second period and 0.5 duty cycle. We calculate the variance sequence frame by frame, either without gradient processing or after applying it, which is shown in Figure \ref{three_pic}. The variance sequence obtained after gradient processing has less noise and can better reflect the vibration patterns.

\begin{figure}[t]
    \vspace{-0.5cm}
    \centering
    \subfloat[][]{\includegraphics[width=0.22\textwidth]{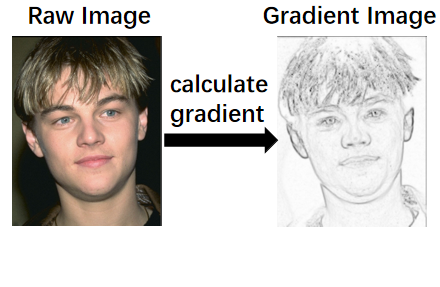}}
    \subfloat[][]{\includegraphics[width=0.26\textwidth]{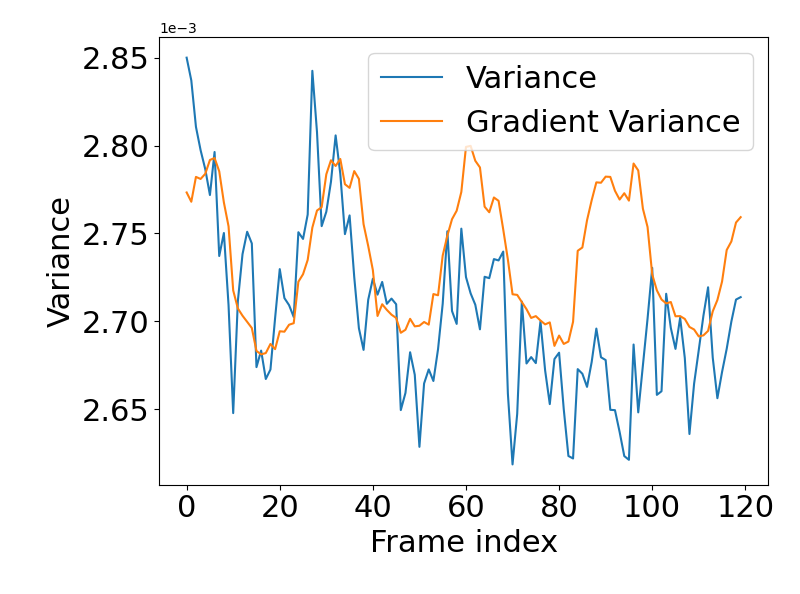}}		
    \caption{(a) The original image and the gradient image. To demonstrate the gradient image more clearly, we invert the image's pixel, meaning that areas with higher pixel values appear darker. (b) We calculate the variance sequence of two videos. One is the raw video without any processing, and another is the video after gradient processing. The variance sequence of the latter one is smoother and more clearly reflects the patterns of the vibration.}
    \label{three_pic}
\end{figure}

\textbf{Landmarks detection}.
Although gradient calculation helps extract the blurriness information, we do not perform it accross the entire image for two reasons. First, computing the gradient requires a lot of computational resources. Second, the entire image may contain some vibration-unrelated changes such as variations of the video's background or alterations of facial expression like blinking and frowning. Therefore, we need to select some areas representing the whole video for subsequent processing. In the previous section, we identify these areas through manual or random selection. In this part, we describe the method for automatically selecting these areas.

Vibration caused blurriness are more likely to occur in edges which are more likely to appear in the face features. Therefore, we use landmark detection algorithm to extract these areas. We make use the Dlib~\cite{dlib} for facial landmark detection. The detection algorithm marks 68 landmarks on the face, which are distributed in a fixed pattern; for instance, points 49 to 68 outline the contour of the mouse, while points 18 to 28 locate the eyebrows. The example of the landmarks is shown in Figure \ref{landmark}(a). As the grayscale value represents the number of edges in this area, which indicates the extent to which vibration affects blurriness, we then calculate the average gradient value within the scope of these rectangles. To avoid the influence of expressions such as blinking and frowning on variance calculation, we exclude the landmarks of eyes and eyebrows from the selection range, and the top n areas with the highest average gradient values in the remaining rectangles are selected to calculate the blurriness degree, where n is a configurable parameter. As the smartphone stays still during detection process, we use the video's first frame to select the areas and keeps their postion in the whole video. 

We demonstrate the effect of selecting areas by landmarks with an experiment. First, in the 4-second video, we randomly pick a 50x50 area for variance sequence calculation. Then, following the algorithm described above, we select one area for variance sequence calculation. The result is shown in Figure \ref{landmark}. Our selected area reflects the vibration pattern better.




\begin{figure}[t]
    \centering
    \vspace{-0.3cm} 

    \subfloat[][]{\includegraphics[width=0.22\textwidth]{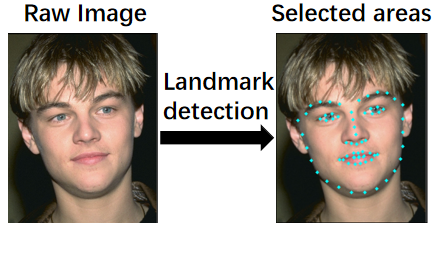}}
    \subfloat[][]{\includegraphics[width=0.26\textwidth]{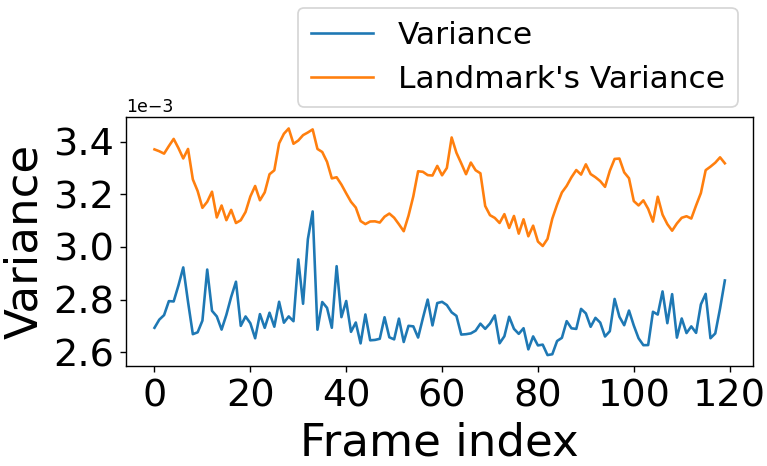}}		
    \caption{(a) The raw image and its facial landmarks. (b) The blue waveform is the variance sequence of the area randomly selected, and the orange waveform is that selected by landmark detection. The latter one better reflects the pattern of the vibration.}
    \label{landmark}
\end{figure}

\begin{figure}[t]
    \centering
    \vspace{-0.3cm} 

    \subfloat[][]{\includegraphics[width=0.24\textwidth]{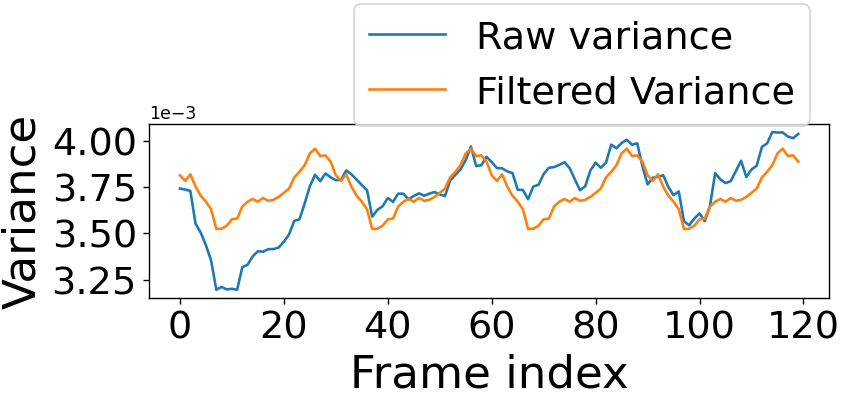}}
    \subfloat[][]{\includegraphics[width=0.24\textwidth]{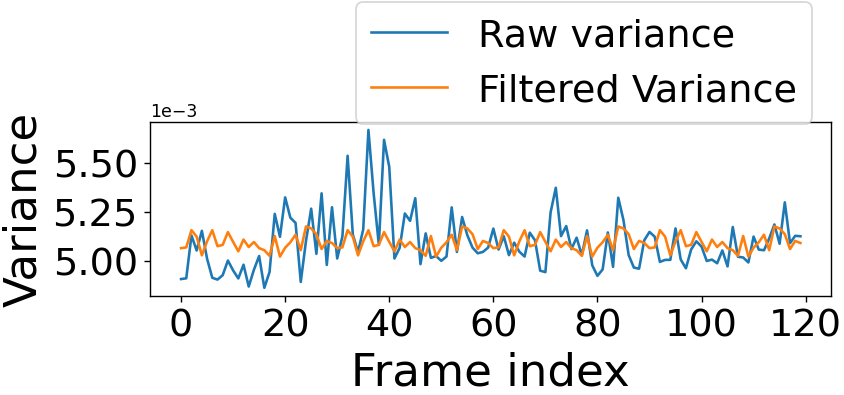}}
    \caption{(a) The raw and filtered variance sequence of the real video. (b) The raw and filtered variance sequence of the fake video.}
    \label{raw_filtered}
\end{figure} 

\begin{table*}[t]
\caption{The detection performance of various methods against different deepfake algorithms.}
\label{tab_detection_performance}
\centering
{\footnotesize
\begin{tblr}{
  width = \linewidth,
  colspec = {Q[106]Q[63]Q[63]Q[63]Q[63]Q[71]Q[71]Q[63]Q[63]Q[63]Q[63]Q[63]Q[63]Q[63]Q[63]},
  cells = {c},
  cell{1}{1} = {r=2}{},
  cell{1}{2} = {c=2}{0.1\linewidth},
  cell{1}{4} = {c=2}{0.1\linewidth},
  cell{1}{6} = {c=2}{0.1\linewidth},
  cell{1}{8} = {c=2}{0.1\linewidth},
  cell{1}{10} = {c=2}{0.1\linewidth},
  cell{1}{12} = {c=2}{0.1\linewidth},
  cell{1}{14} = {c=2}{0.1\linewidth},
  vline{13} = {1}{},
  vline{14} = {1-7}{},
  hline{1,3,8} = {-}{},
}
Methods   & SBI~\cite{shiohara2022detecting} &       & FaceAF~\cite{li2019exposing} &       & CnnDetect~\cite{wang2019cnngenerated} &       & LRNet~\cite{sun2021improving} &       & DFHob~\cite{9428361} &       & Dware~\cite{deepware} &       & Ours  &       \\
          & ACC       & AUC   & ACC          & AUC   & ACC          & AUC   & ACC   & AUC   & ACC       & AUC   & ACC       & AUC   & ACC   & AUC   \\
MFS~                                 & 0.862          & 0.885                                             & 0.625 & 0.655                                                         & 0.784 & 0.843                                                            & 0.848 & 0.891                                                        & 0.766 & 0.792                                                        & 0.832 & 0.873                                                         & \textbf{0.956} & \textbf{0.962}                                     \\
HFace~                               & \textbf{0.964} & \textbf{0.971}                                    & 0.802 & 0.860                                                         & 0.909 & 0.933                                                            & 0.892 & 0.946                                                        & 0.822 & 0.869                                                        & 0.928 & 0.942                                                         & 0.954          & 0.972                                              \\
Fsgan~                               & 0.815          & 0.867                                             & 0.574 & 0.627                                                         & 0.736 & 0.791                                                            & 0.757 & 0.802                                                        & 0.697 & 0.775                                                        & 0.832 & 0.864                                                         & \textbf{0.952} & \textbf{0.968}                                     \\
DFL~                                 & 0.842          & 0.926                                             & 0.677 & 0.713                                                         & 0.763 & 0.785                                                            & 0.826 & 0.853                                                        & 0.822 & 0.878                                                        & 0.742 & 0.796                                                         & \textbf{0.988} & \textbf{0.999}                                     \\
RAI~                                 & 0.820          & 0.841                                             & 0.652 & 0.689                                                         & 0.745 & 0.792                                                            & 0.681 & 0.748                                                        & 0.773 & 0.815                                                        & 0.765 & 0.813                                                         & \textbf{0.968} & \textbf{0.976}                                     \\

\end{tblr}}
\end{table*}

\subsection{Data Processing}
Although at most times, the variance sequence of the selected area can reflect the vibration patterns, it can also be affected by other factors, such as lighting changes and subtle body movements. We capture a 4-second video indoors, deliberately opening the curtains slowly during the recording to introduce subtle lighting variations. The vibration pattern is set with a period of 1 second and a duty cycle of 0.5. The variance sequence of the selected area is shown as the blue waveform in Figure \ref{raw_filtered}(a). The sequence does not reflect the vibration pattern clearly because the lighting variations (and other environmental factors) also influence the variance calculation.

To extract vibration-related components from the variance sequence, we analyze the frequency domain of the \textit{ideal} variance sequence, which is a square wave with a period of 1 second and a duty cycle of 0.5. In theory, the actual variance sequence can be considered as the cumulative effect of various factors influencing the blurriness of the video, whereas the ideal variance sequence can be viewed as the isolated impact of vibration on the blurriness. Therefore, to extract the feature reflecting the vibration patterns, we combine the information of both ideal and actual variance sequences. We perform a Fourier transform on the ideal variance sequence to obtain its frequency spectrum and sort the top 80$\%$ of its non-zero frequency components as the effective ones. We filter out all other frequency components of the actual variance sequence, and the filtered sequence is shown as the orange waveform in Figure \ref{raw_filtered}(a). The sequence drops periodically with a cycle of 30 frames and a duty cycle of about 0.5, which is consistent with the vibration pattern. This way, we have completed the feature extraction, representing the feature as a vector of length 120.

To further demonstrate the effect of the extracted features, we generate a fake video based on the recorded 4-second video. We then apply the same processing to the fake video, with the results illustrated in Figure \ref{raw_filtered}(b). Neither the raw variance sequence nor the filtered one has reflected the vibration pattern. This indicates that our features can differentiate between real and fake videos.

\section{Evaluation}\label{sec_eva}
\subsection{Experiment Setup}\label{sec_eva_setup}
\textbf{Dataset.} Considering the absence of available deepfake datasets including physical probe mechanisms, we use 8 different brands of smartphones to record 15 participants of varying genders and ages to build our own dataset. We place the smartphone on the phone holder 20 cm away from the participant and zoom in twice, aiming at the participant’s face to encompass all his facial features while vibrating the smartphone in different patterns. For phones whose front cameras cannot zoom, we use the rear cameras as a substitute. We record 150 long videos, each 20 seconds in duration. By default, we assume the detection period lasts 4 seconds. We trim 10 clips of 4 seconds long from one long video by randomizing the start time. Therefore, we get a total of 1500 real clips, each 4 seconds long.

Based on the real videos, we use five different deepfake algorithms to generate fake videos: Hififace~\cite{wang2021hififace} and FSGANV2~\cite{nirkin2022fsganv2} which represent the new deepfake algorithm proposed in the academic community, DeepFaceLive~\cite{DeepFaceLive} and RemakerAI~\cite{RemakerAI2023} which represent the widely used online face swap applications, and MobileFaceSwap~\cite{xu2022MobileFaceSwap} which represents the lightweight deepfake algorithm specifically designed for the mobile devices. For each deepfake algorithm, we generate 1500 fake videos, each corresponding to a real video.

\textbf{Ethical Considerations.} We prioritize societal security and ethical concerns. All participants comply with approved IRB protocols, ensuring participant awareness of the usage of their images. Additionally, the deepfake samples created for the study are not utilized beyond its scope and securely discarded post-research.

\textbf{Object of comparison.} We test the performance of 5 existing detection methods from the academic community (SBI~\cite{shiohara2022detecting}, FaceAF~\cite{li2019exposing}, CnnDetect~\cite{wang2019cnngenerated}, LRNet~\cite{sun2021improving}, DFHob~\cite{9428361}), including two of them specifically designed to be lightweight networks~\cite{9428361}~\cite{sun2021improving}. Additionally, we test Deepaware~\cite{deepware}, an online deepfake website, considering its widespread use due to its simplicity and being free of charge. 

\textbf{Classifier design.} We employ a simple two-layer neural network as our classifier with dimensions 120x30 for the hidden layer and 30x1 for the output layer. We utilize ReLU as the activation function and set the learning rate as 0.01. We randomly select 1000 real videos and 1000 fake videos generated solely by DeepFaceLive. For each video, we select 3 areas as the training data. The network converges after 40 epochs. When testing, we also select three areas for every video and determine the authenticity by the majority classfication results among the three areas. It is worth noting that there are possibly better choices for the classifier design, and we will explore the impact of different classifiers on the detection task later.

\textbf{Evaluation metrics and devices.} We use NVIDIA RTX 3060 to build the dataset and implement our detection method. The detection methods run in Ubuntu 22.04.2 LTS. We record the videos using the Xiaomi Redmi 10x, Xiaomi Redmi K50, OPPO Find x6, Huawei Nova9, Xiaomi 14 Ultra, Honor 20, Google Pixel 6a, and Huawei P60. We use Pytorch to reproduce existing detection methods. We utilize standard metrics for assessing the detection methods: the Area Under the Receiver Operating Characteristic Curve (AUC) and Accuracy (ACC). 

\subsection{Overall Performance}\label{sec_eva_op}
In this part, we compare \name\ with other detection models on the dataset to evaluate its capability to classify real and fake videos. Additionally, we measured the computational speed and overhead of various detection methods, demonstrating the efficiency advantages of \name.

\subsubsection{Detection Performance}
Table \ref{tab_detection_performance} shows the results of detecting fake faces generated by various deepfake algorithms by different methods. In all cases, the detection accuracy of \name exceeded 95$\%$. Among the five deepfake algorithms, except for Hififace, \name\ performs better against other deepfake algorithms than the other six detection methods. As our classifier is trained using fake images generated by DeepFaceLive, it reaches the highest accuracy rate of 98.8$\%$ when detecting DeepFaceLive. When facing fake faces generated by RemakerAI, other detection methods perform poorly. We speculate this may be because of the automatic compression of videos when downloading from the internet, resulting in the loss of image details and thereby reducing the detection accuracy. However, this does not affect the detection by \name\, which achieves an accuracy of 96.8$\%$ in detection against RemakerAI.

We use Figure \ref{compare_fig} to explain why \name\ performs well in detecting fake videos. We record a 7-second video and set vibration patterns with a period of 1 second and a duty cycle of 0.5. We use the five deepfake algorithms mentioned above to get the fake videos. We process the six videos and get the variance sequences as shown in Figure \ref{compare_fig}. Notably, the variance sequence drops every 30 frames and has a duty cycle of 50$\%$, which reflects the vibration patterns. In contrast, the variance sequence of the fake videos exhibits irregular noise. This indicates that the deepfake models cannot detect the slight blurriness changes caused by vibrations, resulting in their variance sequence being almost unrelated to the vibration patterns. Such differences between real and fake videos can easily be recognized by a well-designed algorithm.
\begin{figure}[t]
    \centering
    \includegraphics[scale=0.26]{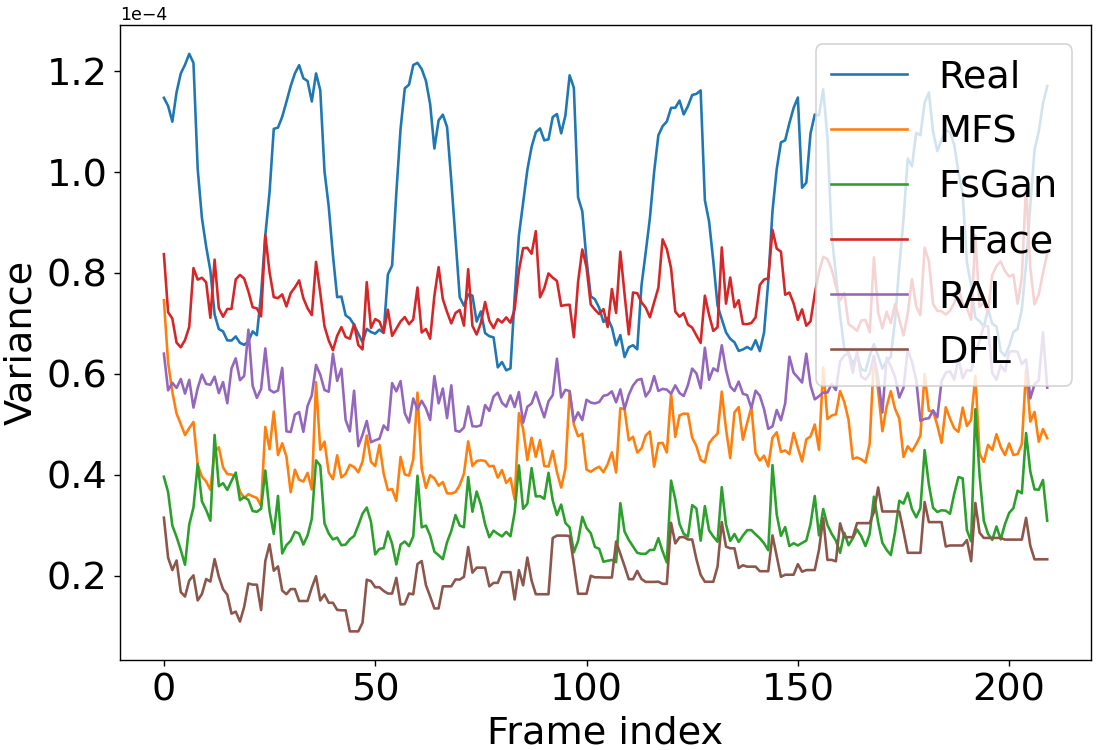}
    \caption{We extract the same 50x50 area from one real video and its corresponding five fake videos to calculate the variance sequences. The waveforms of the real and fake videos are substantially different, which explains why \name\ performs well in detection.}
    \label{compare_fig}
\end{figure}
\subsubsection{Computation Performance}
As \name\ is designed for mobile devices where the computational resource is limited, we discuss the time and memory required by the different detection methods in this part. We use \textit{pmam -x pid} to check the physical memory consumption of the detection process by pid and use the built-in timing function in Python to measure the time consumption. To compare the computational performance, we record a 4-second video and measure the memory and time consumption when detecting it with the 6 detection methods, except for Deepaware, which is deployed online and hides its memory information. 

The result is shown in Figure \ref{comperformance}. Other detection methods require several gigabytes of memory space to load the memory-consumed models. In contrast, \name\ consumes less than 450MB of memory space, which is about one-tenth of the other detection methods. This is because \name\ extracts the features by memory-saving operations like computing the variance, calculating the gradient, performing the Fast Fourier Transform, and using the two-layer neural network for classification. Regarding processing time, \name\ takes 4.52 seconds to complete the processing, which is only 0.56 times the shortest duration among the other methods. The most time-consuming part of \name's detection is calculating the gradient frame by frame within multiple selected areas. However, considering that the variance calculation for each area is independent, we create multiple processes where each process computes the gradient of one area. In this way, calculating the gradient for more areas will not significantly increase the time consumption. 

\begin{figure}[t]
    \centering
    \includegraphics[scale=0.28]{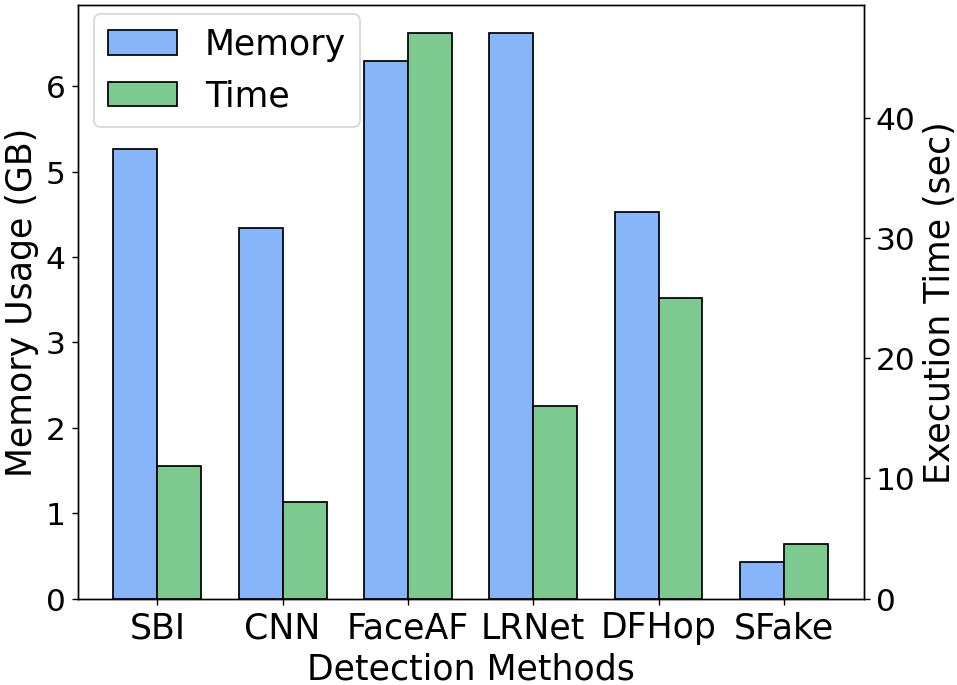}
    \caption{The computation performance of the different detection methods.}
    \label{comperformance}
\end{figure}

\begin{figure*}[t]
    \vspace{-0.6cm}
    \centering
    \subfloat[][]{\includegraphics[width=0.33\textwidth]{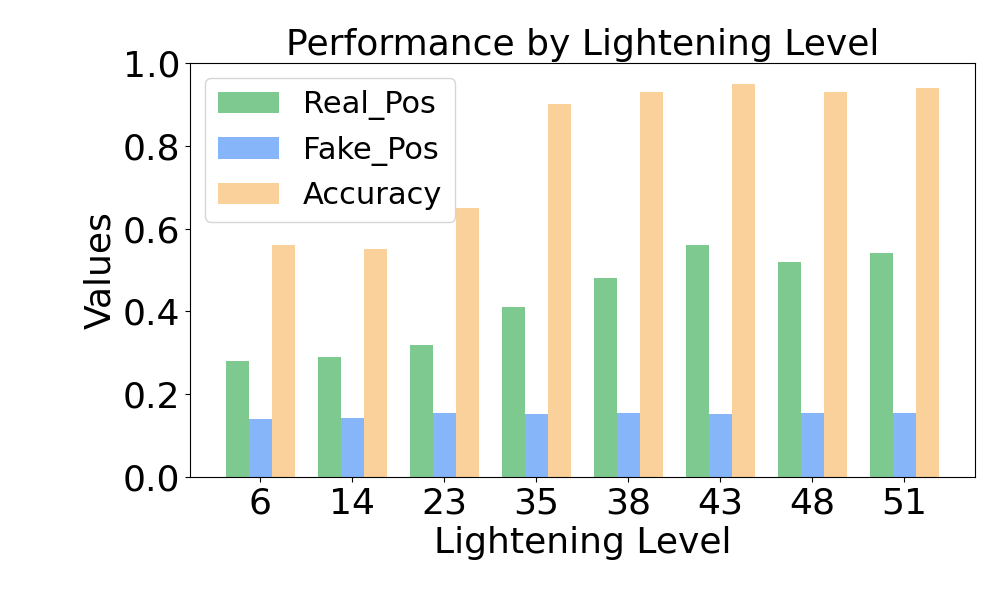}}
    \subfloat[][]{\raisebox{-0.2ex}{\includegraphics[width=0.33\textwidth]{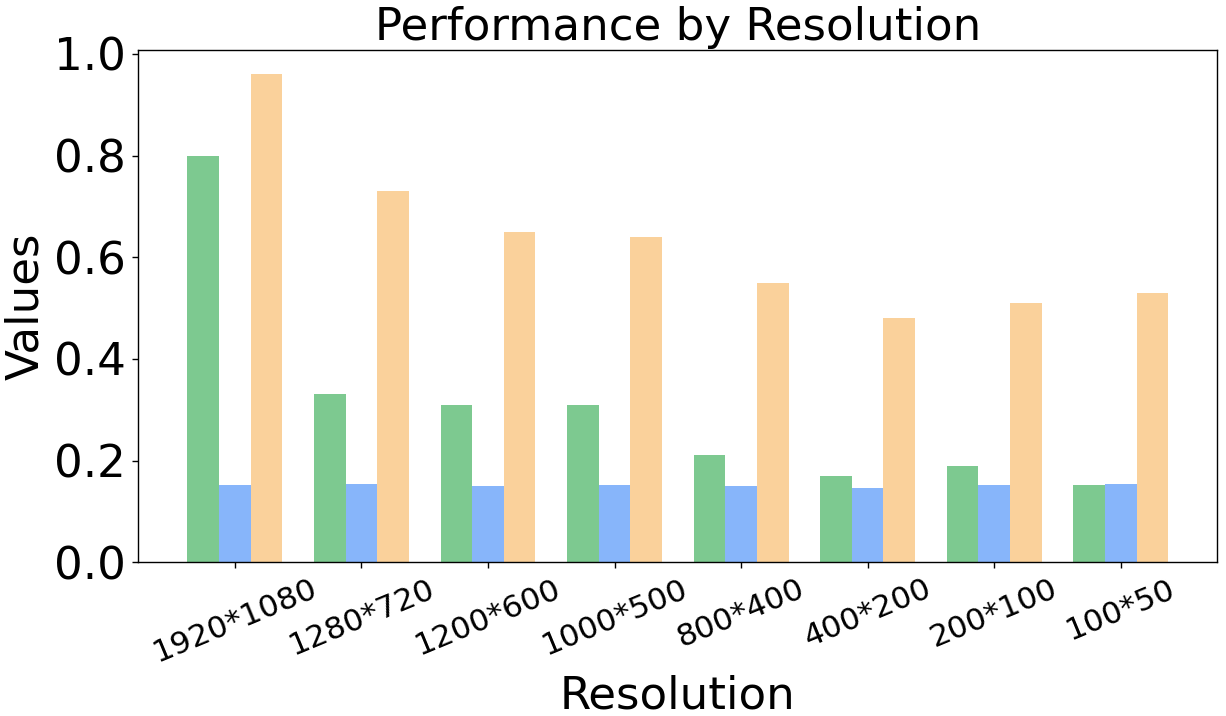}}}\\
    \subfloat[][]{\raisebox{-0.45ex}{\includegraphics[width=0.29\textwidth]{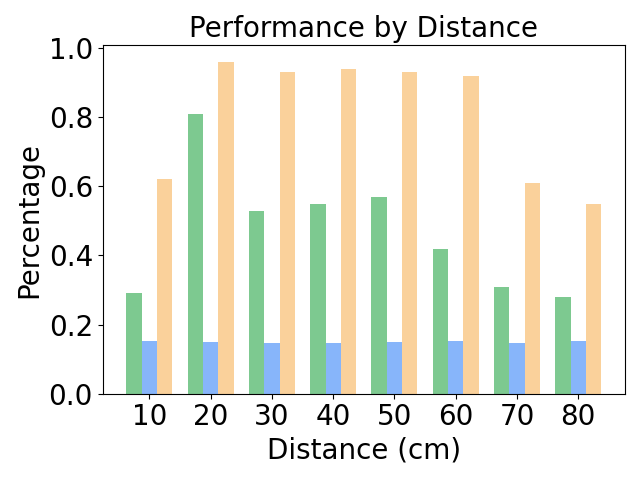}}}	
    \subfloat[][]{\includegraphics[width=0.31\textwidth,height=0.21\textwidth]{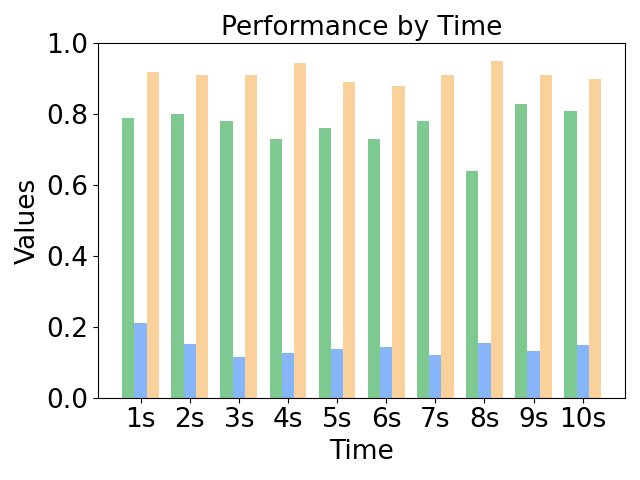}}
    \subfloat[][]{\raisebox{-0.5ex}{\includegraphics[width=0.29\textwidth]{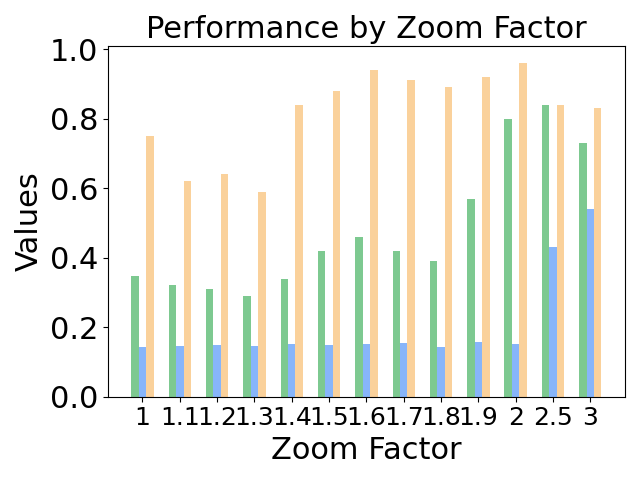}}}	
    \caption{The impact of (a) lightening level, (b) detection time, (c) zoom factor, (d) resolution, and (e) shooting distance on accuracy and POS values for real and fake videos.}
    \label{impactfactor}
\end{figure*}

\subsection{Impact Factors}
Considering the various scenarios of video communications in real life, we explore the relationship between environmental factors and the distinctness of the feature. We launch several experiments to illustrate the relationship between the performance of \name\ and several factors such as lighting conditions, camera resolution, shooting distance, time consumed during the detection process, zoom factor, and classifiers.

\textbf{Metrics and setup.} To measure the distinctness of features, we define the Proportion of Sequence (POS), which means the ratio of the ideal frequency components to all the frequency components in the actual variance sequence. The ideal frequency components are defined as the non-zero frequency bands of the ideal variance sequence. The mathematical expression for POS is
\begin{equation}
    POS = \frac{sum(f_{avs}[where\quad f_{ivs}!=0])}{sum(f_{avs})}
    \label{POS}
\end{equation}
Where $f_{ivs}$ and $f_{avs}$ represent the discrete frequency spectrum of the ideal and the actual variance sequence, respectively. A higher POS value suggests that the vibration more significantly influences the variance sequence. We record a 40-second video for each environmental circumstance and generate the corresponding fake video by RemarkerAI as it presents the most significant challenge for detection models according to our experiment results. Subsequently, we trim 100 video clips of four seconds each from the real and fake videos, choosing the start times randomly. We use POS of real and fake videos and the detection accuracy of the 200 clips as the metrics to determine the distinctness of the feature and the performance of the \name. Unless otherwise specified, all other experimental conditions remain unchanged as illustrated in Section \ref{sec_eva_setup}.

\subsubsection{Lightening Level}
We explore the performance of \name\ under different lighting conditions. We place the smartphone in a small room, adjusting the luminous intensity by controlling the room's lights. We measure the light intensity by the average grayscale value of the video's first frame. The experimental results are shown in Figure \ref{impactfactor}(a). \name\ performs well with a lighting level of 35, akin to illuminating one's face with the light from a computer screen in a dark room. When the lighting level is below 35, the detection accuracy and the real video's POS value decrease rapidly. That may be because insufficient lighting reduces the sharpness of the edges between different color blocks in the footage, resulting in a decrease in the blurriness caused by the vibration. It indicates that \name\ works well with even poor lighting conditions.

\subsubsection{Resolution} 
Considering \name\ may work under poor network conditions which influence the video quality, we explore the detection results at different video resolutions. We first record videos with a resolution of 1920*1080 and then compress the original videos by \textit{cv2.resize} function to generate videos with smaller resolutions. The result is shown in Figure \ref{impactfactor}(b). When the resolution is lower than 1920*1080, the accuracy sharply decreases, along with a decrease in real video POS. That is because the blurring caused by vibration is relatively minor, and reducing the image from high resolution to low resolution is essentially a manual process of blurring the image, which overshadows the blurring caused by vibration, resulting in the decrease of real video POS and accuracy. We find that Tencent Meeting~\cite{TencentMeetings}, Microsoft Teams~\cite{MicrosoftTeamsNetwork}, Zoom~\cite{ZoomSupportArticle}, and Skype~\cite{MicrosoftSkypeResolutions} all support resolutions of 1920*1080 or even higher. Additionally, even if the cloud services for \name\ are inaccessible due to poor network conditions, we can consider deploying \name\ locally. According to the experiment about memory consumption, \name\ consumes less than 450 MB, which can run locally on a mobile device and thereby ensure that the video footage is not affected by the network conditions.

\subsubsection{Shooting distance}
This part explores the relationship between detection performance and shooting distance. We place the smartphone at different distances from the subject's face and record the videos. The result is shown in Figure \ref{impactfactor}(c). When the distance is 20cm, the POS value of the real video reaches its maximum, and the feature is most prominent. Within the range of 20cm to 60cm, the POS value of the real video is significantly greater than that of the fake video, indicating a good detection performance. The detection accuracy decreases when the distance is too close or too far. When the camera is close to the subject, the camera may not encompass the entire facial contour, which reduces the search range when selecting the landmarks with the most gradient value. When the camera is too far from the facial area, the details of the face may not be clear, so vibrations are less likely to affect the variance of the face. As reported, the average distance at which women use mobile devices is 34.7 cm, while for men, it is 38.2 cm~\cite{BOCCARDO2021120}. \name\ performs well in this distance range.

\subsubsection{Detection time} In Section \ref{sec_eva_op}, we set the detection time as 4 seconds, which indicates the classifier's input should be a vector with a length of 120 (as the frame rate is 30 fps). To evaluate different lengths of detection time, we trim or splice the existing variance segmentations to match the length of the input and retrain the model. The results shown in Figure \ref{impactfactor}(d) indicate that, apart from 4s or 8s, the accuracy decreases to varying degrees at other detection times. Notably, for 5s or 6s, the accuracy drops below 90$\%$, possibly due to a decline in the quality of the training data. The accuracy at 8s detection time is relatively unaffected because the duration time of 8 seconds is an integer multiple of the original 4s, and their frequency components share similarities. We re-trim 200 videos to lengths of 5 seconds and 6 seconds from the original dataset and retrain two models with input sizes of 150 and 180 (the frame numbers of 5 and 6 seconds), respectively. We find the accuracy return to 94$\%$ and 95$\%$, proving that the detection time can not significantly affect the accuracy. This is because when the user remains stationary, as long as the detection period exceeds the vibration cycle, the variance sequence does not change significantly in the subsequent time.

\subsubsection{Zoom factor} When the zoom factor increases, the focal length increases, and according to Eq. \ref{L1} and Eq. \ref{L2}, the degree of blurriness caused by vibration also increases. We explore the impact of different focal lengths on detection performance, and the result is shown in Figure \ref{impactfactor}(e). The optimal range for the zoom factor is from 1.6 to 2. Within this range, the recognition accuracy is above 89$\%$. When the zoom factor is too small, the vibration cannot cause significant variance changes in the image, resulting in poor accuracy. When the zoom factor is too significant, we find that the POS value of the fake videos unexpectedly increases. The variance of the fake videos being affected by vibration means the deepfake algorithm also has a specific adaptability to blurriness: when the real video gets too blurred, the corresponding fake video also becomes blurred. Based on our testing, apart from DeepFaceLive and RemarkerAI, other models do not have similar effects. Even if RemakerAI and DeepFaceLive can somehow track the blurriness changes of the real videos, the variance sequence of the fake video still cannot reflect vibration patterns: \name\ maintains the recognition accuracy at 84$\%$ and 83$\%$ when the magnification factor is increased to 2.5 and 3, respectively.

\subsubsection{Classifier}
As mentioned earlier, the classifier is not the primary determinant of the \name\ detection performance. In our design, \name\ randomly selects classifiers to increase the uncertainty of the detection system, thereby enhancing the difficulty for attackers to breach the defense. However, different classifiers may exhibit slight variations in the task. In this part, we compare the detection performance of five simple binary classifiers. As the fake videos in the dataset are generated only by DeepFaceLive, We separately test the classifiers' accuracy on the fake videos solely generated by DeepFaceLive (referred to as Accuracy1) and on the fake videos generated by multiple deepfake algorithms (referred to as Accuracy2). The result is shown in Table \ref{tab_classifier}. We test five classification models: Logistic Regression, a two-layer Neural Network, K-Nearest Neighbors (KNN), Support Vector Machine (SVM), and Classification And Regression Tree (CART). All five classifiers can effectively distinguish fake videos generated by DeepFakeLive. The two-layer Neural Network and KNN perform best with an accuracy of 98.81$\%$ and 98.33$\%$, respectively. When we select the RBF kernel function with a degree of 5, the SVM's accuracy reaches 94.75$\%$. When using these classifiers to detect fake videos generated by other deepfake algorithms, accuracy decreases to varying degrees. The CART model suffers the most pronounced decline, possibly due to overfitting caused by our selection of an unsuitable depth. The two-layer neural network and KNN demonstrate better generalization abilities, with current accuracies at 97.25$\%$ and 91.23$\%$, respectively. The experiments prove that Logistic Regression, two-layer Neural Networks, KNN, and SVM can all serve as potential classifier alternatives. However, this does not imply that these classifiers are the best choices. \name\ can be paired with more complex classifiers to increase the overall defense's uncertainty in detection, thereby making it harder for attackers to breach our defense.

\definecolor{MineShaft}{rgb}{0.2,0.2,0.2}
\definecolor{Shark}{rgb}{0.125,0.129,0.141}
\begin{table}
\centering
\caption{The detection performance of different classifier.}
\label{tab_classifier}
{\footnotesize
    \begin{tblr}{
      width = \linewidth,
      colspec = {Q[202]Q[156]Q[137]Q[137]Q[137]Q[137]},
      cells = {c},
      cell{1}{3} = {fg=MineShaft},
      cell{1}{6} = {fg=Shark},
      hline{1-2,4} = {-}{},
    }
    Classifier & Logistic & NN     & KNN    & SVM    & CART   \\
    Accuracy1  & 0.9224   & 0.9881 & 0.9833 & 0.9475 & 0.9058 \\
    Accuracy2  & 0.8872   & 0.9725 & 0.9123 & 0.8613   & 0.7624 
    \end{tblr}
}
\end{table}

\subsection{Discussion}
In our previous experiments, we place the phone on a phone holder to prevent hand movement and affect the footage. However, in most cases, users hold the phone to complete the detection process. With the natural tremor of the hands, the smartphone slightly but rapidly moves parallel to the plane of the phone, thereby impacting the recognition by \name. We record ten videos of 40 seconds with a hand holding the smartphone. The vibration period is 1 second, and the duty cycle is 0.5. We slice it into one hundred 4-second clips. We randomly select one clip to calculate the raw and filtered variance sequence without other processes. The waveform reflects the vibration patterns but not so clearly, as shown in Figure \ref{diff_process}(a). The POS of the real video decreases to 0.25, and the accuracy decreases to 58$\%$, significantly impacting performance. To solve this problem and maintain the performance, we provide three potential solutions.

\begin{figure}[t]
    \centering
    \vspace{-0.3cm}

    \subfloat[][]{\raisebox{1.3ex}{\includegraphics[width=0.23\textwidth]{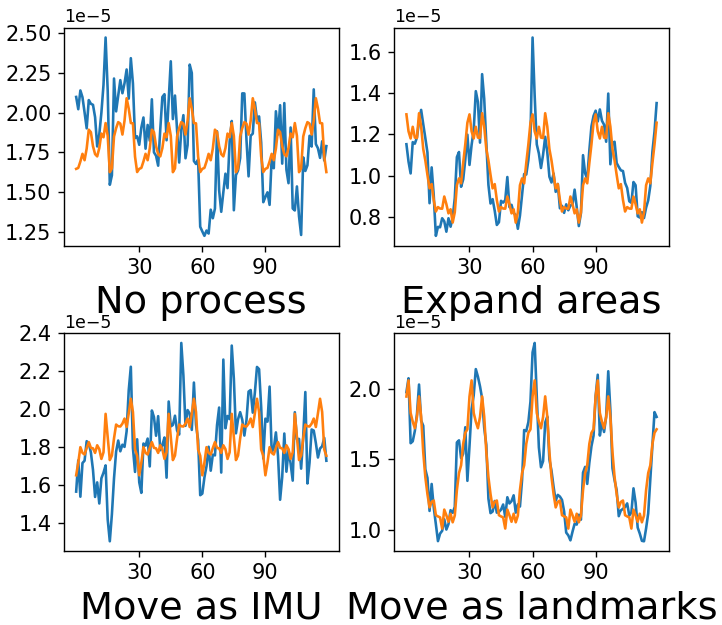}}}
    \subfloat[][]{\includegraphics[width=0.26\textwidth]{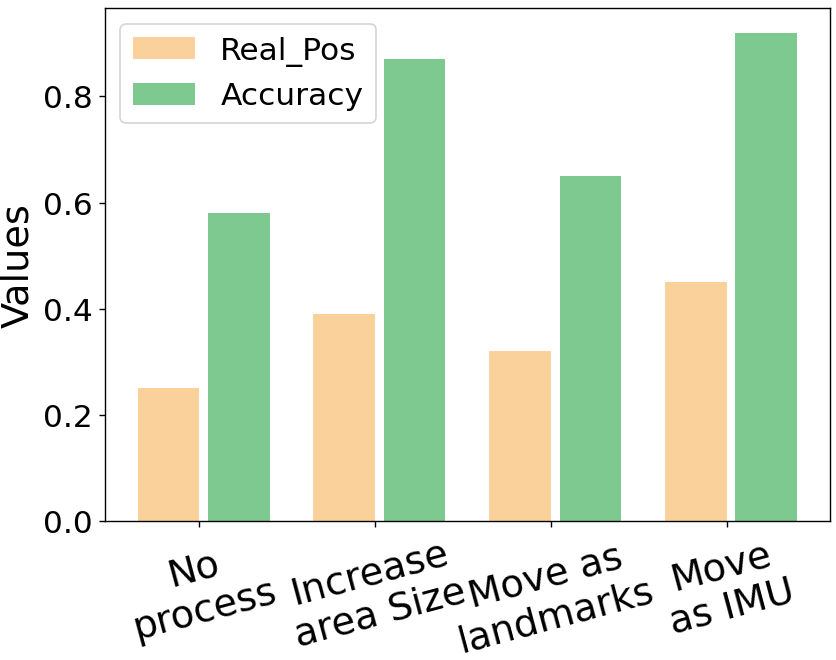}}\\
    \caption{(a) The raw and filtered variance sequences under different process methods with the hand holding the smartphone. (b) The POS of the real videos under different processes.}
    \label{diff_process}
\end{figure}

\textbf{Expand size of selected area.} 
We set the default selected area size to 50 in our previous experiments, which is very small and can accurately capture the blurriness caused by vibrations when the footage is still. However, due to its small size, even slight movements can cause significant changes in the pixel values within the selected area. When the footage moves naturally with the person's hands, the variance is affected not only by the phone's vibration but also by the variations of image content in the selected area. To address this issue, we increase the selected area size to 300. This reduces the proportion of content within the selected area that changes due to hand movement. Experimental results indicate that by expanding the selected area, we have increased the POS value to 0.39 and achieved an accuracy rate of 87$\%$.

\textbf{Move the selected area by IMU data.} The relative movement between the camera and the face impacts the variance sequence calculations. If we can calculate how much distance in pixels the whole image has shifted due to the camera movement, we can then move the selected area in the same direction and distance to frame the same part of the footage. Using the principle of similar triangles, we estimate the number of pixels by which the image moved as Eq. \ref{imu}
\begin{equation}
    p = \frac{d}{D}*f
    \label{imu}
\end{equation}
Where $p$ is the number of pixels the image moves, $d$ is the distance the camera moves, $D$ is the object distance, and $f$ is the focus length. We assume the object distance is 20 cm. We measure the distance of smartphone movement by IMU data according to Eq. \ref{acc_dis}, check the datasheet for focus length, and calculate the number of pixels the whole footage moved. Subsequently, we move the selected area with the same number of pixels in the same direction. The result shows that we marginally improve the performance of \name. We increase the POS to 0.32 and the accuracy of detection to 65$\%$. We believe that this is because of inaccuracy in estimating the distance from the face to the camera, and the IMU errors also limit performance improvement.

\textbf{Move the selected area by centroid of landmarks.} We perform landmark detection frame by frame on the video. As the positions of facial features may change due to expressions such as blinking or frowning, whereas the contour of a person's face tends to be more fixed, we characterize the position of the scene by calculating the centroid of landmarks along the facial contour. We use the offset of the centroid position to represent the overall shift of the smartphone within the frame. We accordingly adjust the position of the selected area so that each frame's selected area gets the same content. The result shows we improve the POS to 0.45 and the accuracy to 92$\%$.

\section{Related Work}\label{sec_rw}
In this part, we briefly review the recent work on deepfake detection. Existing deepfake detection methods mainly discover the inconsistencies in the footage across spatial, temporal, and frequency domains. For example, spatially, one can detect face swaps by examining if the distribution of features such as color~\cite{he2019detection}~\cite{abdullah2023deepfake}, noise~\cite{zhou2017two}~\cite{nguyen2019capsule}~\cite{wang2023noise}, and gray level~\cite{pishori2020detecting}~\cite{son2021measurement} are consistent throughout the image. Temporarily, physiological signals like the frequency of blinking~\cite{li2018ictu} and noding~\cite{yang2019exposing}, gaze angles~\cite{peng2024deepfakes}, heart rate~\cite{hernandez2020deepfakeson} and audio-visual modal inconsistency~\cite{cozzolino2023audio}~\cite{feng2023self}~\cite{haliassos2022leveraging}~\cite{yang2023avoid} are widely used for deepfake detection. In the frequency domain, Qian et al.~\cite{qian2020thinking} propose a novel network leveraging frequency-aware features and local frequency statistics through a two-stream collaborative learning framework; Li et al.~\cite{Li_2021_CVPR} introduce a frequency-aware discriminative feature learning framework that uses single-center loss to improve class separation; Miao et al.~\cite{9854878} leverages a dual-branch structure combining CNNs and transformers to pinpoint frequency-domain flaws in forged faces effectively; Guo et al.~\cite{10286083} introduces Space-Frequency Interactive Convolution (SFIConv), featuring a Multichannel Constrained Separable Convolution (MCSConv) to capture high-frequency tampering traces left by Deepfake.

Contrary to passively recognizing the video's features, our work focuses on actively sending probes to attackers' smartphones to introduce readily recognizable features into videos. By setting the patterns and timing of introducing features, the defender can break free from constantly evolving attack methods and take the initiative in detection without excessively investigating the means and traces of how deepfake creates counterfeit videos.

\section{Conclusion}\label{sec_con}
In this work, we propose \name, a new deepfake detection method that actively introduces features into the video footage by physical probes. We explore the phenomenon of smartphone vibration, video blurriness, and feature extraction, on the base of which we design and implement \name. To test it, we build up a dataset with 8 brands of smartphones, 15 participants, and 5 existing deepfake algorithms. We evaluate \name\ and compare it with 6 other existing detection methods, and the result shows \name\ has a higher detection accuracy with less memory consumed and faster processing speed.
\ifCLASSOPTIONcompsoc
  \section*{Acknowledgments}
\else
  \section*{Acknowledgment}
\fi

\bibliographystyle{plain}

\bibliography{ref}

\begin{thebibliography}{10}

\bibitem{RemakerAI2023}
Ai face swap online.
\newblock \url{https://remaker.ai/en}.
\newblock Accessed: 2023-04-28.

\bibitem{ZoomSupportArticle}
Getting started with zoom.
\newblock Zoom Help Center, 2023.
\newblock Accessed: 2023-05-31.

\bibitem{MicrosoftTeamsNetwork}
Prepare your organization's network for microsoft teams.
\newblock Microsoft Docs, 2023.
\newblock Accessed: 2023-05-31.

\bibitem{TencentMeetings}
Tencent meeting help center.
\newblock Tencent Meetings Support, 2023.
\newblock Accessed: 2023-05-31.

\bibitem{MicrosoftSkypeResolutions}
Video: Video resolutions in skype for business.
\newblock Microsoft Docs, 2023.
\newblock Accessed: 2023-05-31.

\bibitem{abdullah2023deepfake}
Mohammed~Thajeel Abdullah and Nada Hussein~M Ali.
\newblock Deepfake detection improvement for images based on a proposed method for local binary pattern of the multiple-channel color space.
\newblock {\em International Journal of Intelligent Engineering \& Systems}, 16(3), 2023.

\bibitem{agarwal2019protecting}
Shruti Agarwal, Hany Farid, Yuming Gu, Mingming He, Koki Nagano, and Hao Li.
\newblock Protecting world leaders against deep fakes.
\newblock In {\em Proc. of the 32nd IEEE/CVF CVPRW}, volume~1, page~38, 2019.

\bibitem{news4}
JON BATEMAN.
\newblock Get ready for deepfakes to be used in financial scams.
\newblock Online, 2020-08-10.
\newblock \url{https://carnegieendowment.org/2020/08/10/get-ready-for-deepfakes-to-be-used-in-financial-scams-pub-82469}.

\bibitem{BOCCARDO2021120}
Laura Boccardo.
\newblock Viewing distance of smartphones in presbyopic and non-presbyopic age.
\newblock {\em Journal of Optometry}, 14(2):120--126, 2021.

\bibitem{chen2023watching}
Han Chen, Yuezun Li, Dongdong Lin, Bin Li, and Junqiang Wu.
\newblock Watching the big artifacts: Exposing deepfake videos via bi-granularity artifacts.
\newblock {\em Pattern Recognition}, 135:109179, 2023.

\bibitem{9428361}
Hong-Shuo Chen, Mozhdeh Rouhsedaghat, Hamza Ghani, Shuowen Hu, Suya You, and C.-C. Jay~Kuo.
\newblock Defakehop: A light-weight high-performance deepfake detector.
\newblock In {\em Proc. of the 21st IEEE ICME}, pages 1--6, 2021.

\bibitem{chen2022ost}
Liang Chen, Yong Zhang, Yibing Song, Jue Wang, and Lingqiao Liu.
\newblock Ost: Improving generalization of deepfake detection via one-shot test-time training.
\newblock {\em Advances in Neural Information Processing Systems}, 35:24597--24610, 2022.

\bibitem{chhabra2023low}
Saheb Chhabra, Kartik Thakral, Surbhi Mittal, Mayank Vatsa, and Richa Singh.
\newblock Low quality deepfake detection via unseen artifacts.
\newblock {\em IEEE Trans. on Artificial Intelligence}, 2023.

\bibitem{conotter2014physiologically}
Valentina Conotter, Ecaterina Bodnari, Giulia Boato, and Hany Farid.
\newblock Physiologically-based detection of computer generated faces in video.
\newblock In {\em Proc. of the 21st IEEE ICIP}, pages 248--252. IEEE, 2014.

\bibitem{cozzolino2023audio}
Davide Cozzolino, Alessandro Pianese, Matthias Nie{\ss}ner, and Luisa Verdoliva.
\newblock Audio-visual person-of-interest deepfake detection.
\newblock In {\em Proc. of the 32nd IEEE/CVF CVPR}, pages 943--952, 2023.

\bibitem{deepware}
{Deepware AI}.
\newblock Deepware ai: Deepfake detection solutions.
\newblock \url{https://deepware.ai/}, 2023.
\newblock Accessed: 2023-05-29.

\bibitem{dhanaraj2024face}
Rachel DHANARAJ and M~Sridevi.
\newblock Face warping deepfake detection and localization in a digital video using transfer learning approach.
\newblock {\em Journal of Metaverse}, 4(1):11--20, 2024.

\bibitem{dolhansky2020deepfake}
Brian Dolhansky, Joanna Bitton, Ben Pflaum, Jikuo Lu, Russ Howes, Menglin Wang, and Cristian~Canton Ferrer.
\newblock The deepfake detection challenge (dfdc) dataset.
\newblock {\em arXiv preprint arXiv:2006.07397}, 2020.

\bibitem{feng2023self}
Chao Feng, Ziyang Chen, and Andrew Owens.
\newblock Self-supervised video forensics by audio-visual anomaly detection.
\newblock In {\em Proc. of the 36th IEEE/CVF CVPR}, pages 10491--10503, 2023.

\bibitem{GSMArena2020XiaomiRedmi10X}
GSMArena.
\newblock Xiaomi redmi 10x 4g.
\newblock \url{https://www.gsmarena.com/xiaomi_redmi_10x_4g-10202.php}, May 2020.
\newblock Accessed: 2023-05-12.

\bibitem{10286083}
Zhiqing Guo, Zhenhong Jia, Liejun Wang, Dewang Wang, Gaobo Yang, and Nikola Kasabov.
\newblock Constructing new backbone networks via space-frequency interactive convolution for deepfake detection.
\newblock {\em IEEE Trans. on Information Forensics and Security}, 19:401--413, 2024.

\bibitem{haliassos2022leveraging}
Alexandros Haliassos, Rodrigo Mira, Stavros Petridis, and Maja Pantic.
\newblock Leveraging real talking faces via self-supervision for robust forgery detection.
\newblock In {\em Proc. of the 35th IEEE/CVF CVPR}, pages 14950--14962, 2022.

\bibitem{he2019detection}
Peisong He, Haoliang Li, and Hongxia Wang.
\newblock Detection of fake images via the ensemble of deep representations from multi color spaces.
\newblock In {\em Proc. of the 26th IEEE ICIP}, pages 2299--2303. IEEE, 2019.

\bibitem{he2021forgerynet}
Yinan He, Bei Gan, Siyu Chen, Yichun Zhou, Guojun Yin, Luchuan Song, Lu~Sheng, Jing Shao, and Ziwei Liu.
\newblock Forgerynet: A versatile benchmark for comprehensive forgery analysis.
\newblock In {\em Proc. of the 34th IEEE/CVF CVPR}, pages 4360--4369, 2021.

\bibitem{news1}
Kathleen~Magramo Heather~Chen.
\newblock Finance worker pays out \textdollar 25 million after video call with deepfake ‘chief financial officer’.
\newblock Online, 2024-02-04.
\newblock \url{https://edition.cnn.com/2024/02/04/asia/deepfake-cfo-scam-hong-kong-intl-hnk/index.html}.

\bibitem{hernandez2020deepfakeson}
Javier Hernandez-Ortega, Ruben Tolosana, Julian Fierrez, and Aythami Morales.
\newblock Deepfakeson-phys: Deepfakes detection based on heart rate estimation.
\newblock {\em arXiv preprint arXiv:2010.00400}, 2020.

\bibitem{hsu2021deepfake}
Hsuan-Wei Hsu and Jian-Jiun Ding.
\newblock Deepfake algorithm using multiple noise modalities with two-branch prediction network.
\newblock In {\em Proc. of the 13rd IEEE APSIPA ASC}, pages 1662--1669, 2021.

\bibitem{DeepFaceLive}
IPerov.
\newblock Deepfacelive.
\newblock \url{https://github.com/iperov/DeepFaceLive}, 2024.
\newblock GitHub repository.

\bibitem{jiang2020deeperforensics}
Liming Jiang, Ren Li, Wayne Wu, Chen Qian, and Chen~Change Loy.
\newblock Deeperforensics-1.0: A large-scale dataset for real-world face forgery detection.
\newblock In {\em Proc. of the 33rd IEEE/CVF CVPR}, pages 2889--2898, 2020.

\bibitem{kim2021fretal}
Minha Kim, Shahroz Tariq, and Simon~S Woo.
\newblock Fretal: Generalizing deepfake detection using knowledge distillation and representation learning.
\newblock In {\em Proc. of the 34th IEEE/CVF CVPR}, pages 1001--1012, 2021.

\bibitem{dlib}
Davis King.
\newblock dlib.
\newblock \url{https://github.com/davisking/dlib}, Access year.

\bibitem{korshunov2018speaker}
Pavel Korshunov and S{\'e}bastien Marcel.
\newblock Speaker inconsistency detection in tampered video.
\newblock In {\em Proc. of the 26th EUSIPCO}, pages 2375--2379. IEEE, 2018.

\bibitem{Li_2021_CVPR}
Jiaming Li, Hongtao Xie, Jiahong Li, Zhongyuan Wang, and Yongdong Zhang.
\newblock Frequency-aware discriminative feature learning supervised by single-center loss for face forgery detection.
\newblock In {\em Proc. of the 34th IEEE/CVF CVPR}, pages 6458--6467, June 2021.

\bibitem{li2022artifacts}
Xin Li, Rongrong Ni, Pengpeng Yang, Zhiqiang Fu, and Yao Zhao.
\newblock Artifacts-disentangled adversarial learning for deepfake detection.
\newblock {\em IEEE Trans. on Circuits and Systems for Video Technology}, 33(4):1658--1670, 2022.

\bibitem{li2018ictu}
Yuezun Li, Ming-Ching Chang, and Siwei Lyu.
\newblock In ictu oculi: Exposing ai generated fake face videos by detecting eye blinking.
\newblock {\em arXiv preprint arXiv:1806.02877}, 2018.

\bibitem{li2018exposing}
Yuezun Li and Siwei Lyu.
\newblock Exposing deepfake videos by detecting face warping artifacts.
\newblock {\em arXiv preprint arXiv:1811.00656}, 2018.

\bibitem{li2019exposing}
Yuezun Li and Siwei Lyu.
\newblock Exposing deepfake videos by detecting face warping artifacts.
\newblock In {\em Proc. of the 32nd IEEE/CVF CVPRW}, 2019.

\bibitem{li2019celeb}
Yuezun Li, Xin Yang, Pu~Sun, Honggang Qi, and Siwei Lyu.
\newblock Celeb-df (v2): A new dataset for deepfake forensics.
\newblock {\em arXiv preprint arXiv}, 2019.

\bibitem{li2020celeb}
Yuezun Li, Xin Yang, Pu~Sun, Honggang Qi, and Siwei Lyu.
\newblock Celeb-df: A large-scale challenging dataset for deepfake forensics.
\newblock In {\em Proc. of the 33th IEEE/CVF CVPR}, pages 3207--3216, 2020.

\bibitem{liu2024enhancing}
Qingtong Liu, Ziyu Xue, Haitao Liu, and Jing Liu.
\newblock Enhancing deepfake detection with diversified self-blending images and residuals.
\newblock {\em IEEE Access}, 2024.

\bibitem{long2023side}
Yan Long, Pirouz Naghavi, Blas Kojusner, Kevin Butler, Sara Rampazzi, and Kevin Fu.
\newblock {Side Eye: Characterizing the Limits of POV Acoustic Eavesdropping from Smartphone Cameras with Rolling Shutters and Movable Lenses}.
\newblock In {\em Proc. of the 43rd IEEE S\&P}, pages 1857--1874, 2023.

\bibitem{news5}
Aqil~Haziq Mahmud.
\newblock Deep dive into deepfakes: Frighteningly real and sometimes used for the wrong things.
\newblock Online, 2021-10-22.
\newblock \url{https://www.channelnewsasia.com/singapore/deepfakes-ai-security-threat-face-swapping-2252161}.

\bibitem{marra2019gans}
Francesco Marra, Diego Gragnaniello, Luisa Verdoliva, and Giovanni Poggi.
\newblock Do gans leave artificial fingerprints?
\newblock In {\em Proc. of the 2nd IEEE MIPR}, pages 506--511. IEEE, 2019.

\bibitem{masi2020two}
Iacopo Masi, Aditya Killekar, Royston~Marian Mascarenhas, Shenoy~Pratik Gurudatt, and Wael AbdAlmageed.
\newblock Two-branch recurrent network for isolating deepfakes in videos.
\newblock In {\em Proc. of the 16th ECCV}, pages 667--684. Springer, 2020.

\bibitem{9854878}
Changtao Miao, Zichang Tan, Qi~Chu, Nenghai Yu, and Guodong Guo.
\newblock Hierarchical frequency-assisted interactive networks for face manipulation detection.
\newblock {\em IEEE Trans. on Information Forensics and Security}, 17:3008--3021, 2022.

\bibitem{mirsky2021creation}
Yisroel Mirsky and Wenke Lee.
\newblock The creation and detection of deepfakes: A survey.
\newblock {\em ACM computing surveys (CSUR)}, 54(1):1--41, 2021.

\bibitem{mutillo2023knowledge}
V{\'\i}ctor Mutillo-Ligorred, Irene Covaleda, Nora Ramos-Vallecillo, and Leticia Fayos.
\newblock Knowledge, integration and scope of deepfakes in arts dducation: The development of critical thinking in postgraduate students in primary education and master’s degree in secondary education.
\newblock Technical report, 2023.

\bibitem{nadimpalli2022improving}
Aakash~Varma Nadimpalli and Ajita Rattani.
\newblock On improving cross-dataset generalization of deepfake detectors.
\newblock In {\em Proc. of the 35th IEEE/CVF CVPR}, pages 91--99, 2022.

\bibitem{news6}
CCTV news client.
\newblock Quality report of the week | “ai face-changing” more scams! how do we prevent that?
\newblock Online, 2024-03-17.
\newblock \url{https://content-static.cctvnews.cctv.com/snow-book/index.html?item_id=1229629131018651468&t=1710659368802&toc_style_id=feeds_default&track_id=160F13C3-28DD-496B-BAE8-3B2C2B5D12ED_732352771696&share_to=wechat}.

\bibitem{nguyen2019capsule}
Huy~H Nguyen, Junichi Yamagishi, and Isao Echizen.
\newblock Capsule-forensics: Using capsule networks to detect forged images and videos.
\newblock In {\em Proc. of the 44th IEEE ICASSP}, pages 2307--2311. IEEE, 2019.

\bibitem{nirkin2022fsganv2}
Yuval Nirkin, Yosi Keller, and Tal Hassner.
\newblock {FSGANv2}: Improved subject agnostic face swapping and reenactment.
\newblock {\em IEEE Transactions on Pattern Analysis and Machine Intelligence}, 2022.

\bibitem{openstax2016collegephysics}
OpenStax.
\newblock College physics, 2016.

\bibitem{pei2024deepfake}
Gan Pei, Jiangning Zhang, Menghan Hu, Guangtao Zhai, Chengjie Wang, Zhenyu Zhang, Jian Yang, Chunhua Shen, and Dacheng Tao.
\newblock Deepfake generation and detection: A benchmark and survey.
\newblock {\em arXiv preprint arXiv:2403.17881}, 2024.

\bibitem{peng2024deepfakes}
Chunlei Peng, Zimin Miao, Decheng Liu, Nannan Wang, Ruimin Hu, and Xinbo Gao.
\newblock Where deepfakes gaze at? spatial-temporal gaze inconsistency analysis for video face forgery detection.
\newblock {\em IEEE Trans. on Information Forensics and Security}, 2024.

\bibitem{pishori2020detecting}
Armaan Pishori, Brittany Rollins, Nicolas van Houten, Nisha Chatwani, and Omar Uraimov.
\newblock Detecting deepfake videos: An analysis of three techniques.
\newblock {\em arXiv preprint arXiv:2007.08517}, 2020.

\bibitem{qian2020thinking}
Yuyang Qian, Guojun Yin, Lu~Sheng, Zixuan Chen, and Jing Shao.
\newblock Thinking in frequency: Face forgery detection by mining frequency-aware clues.
\newblock In {\em Proc. of the 16th ECCV}, pages 86--103. Springer, 2020.

\bibitem{news2}
Reuters.
\newblock 'deepfake' scam in china fans worries over ai-driven fraud.
\newblock Online, 2023-05-22.
\newblock \url{https://www.reuters.com/technology/deepfake-scam-china-fans-worries-over-ai-driven-fraud-2023-05-22/}.

\bibitem{rossler2019faceforensics++}
Andreas Rossler, Davide Cozzolino, Luisa Verdoliva, Christian Riess, Justus Thies, and Matthias Nie{\ss}ner.
\newblock Faceforensics++: Learning to detect manipulated facial images.
\newblock In {\em Proc. of the 32nd IEEE/CVF CVPR}, pages 1--11, 2019.

\bibitem{shiohara2022detecting}
Kaede Shiohara and Toshihiko Yamasaki.
\newblock Detecting deepfakes with self-blended images.
\newblock In {\em Proc. of the 35th IEEE/CVF CVPR}, pages 18720--18729, 2022.

\bibitem{son2021measurement}
Seok~Bin Son, Seong~Hee Park, and Youn~Kyu Lee.
\newblock A measurement study on gray channel-based deepfake detection.
\newblock In {\em Proc. of the 2nd IEEE ICTC}, pages 428--430. IEEE, 2021.

\bibitem{straub2019using}
Jeremy Straub.
\newblock Using subject face brightness assessment to detect ‘deep fakes’(conference presentation).
\newblock In {\em Real-Time Image Processing and Deep Learning 2019}, volume 10996, page 109960H. SPIE, 2019.

\bibitem{sun2021improving}
Zekun Sun, Yujie Han, Zeyu Hua, Na~Ruan, and Weijia Jia.
\newblock Improving the efficiency and robustness of deepfakes detection through precise geometric features.
\newblock In {\em Proc. of the 34th IEEE/CVF CVPR}, pages 3609--3618, 2021.

\bibitem{usukhbayar2020deepfake}
Binderiya Usukhbayar and Sean Homer.
\newblock Deepfake videos: The future of entertainment.
\newblock {\em Research Gate: Berlin, Germany}, 2020.

\bibitem{vairamani2022analyzing}
Ajantha~Devi Vairamani.
\newblock Analyzing deepfakes videos by face warping artifacts.
\newblock In {\em DeepFakes}, pages 35--55. CRC Press, 2022.

\bibitem{wang2019cnngenerated}
Sheng-Yu Wang, Oliver Wang, Richard Zhang, Andrew Owens, and Alexei~A Efros.
\newblock Cnn-generated images are surprisingly easy to spot...for now.
\newblock In {\em Proc. of the 33rd IEEE/CVF CVPR}, 2020.

\bibitem{wang2023noise}
Tianyi Wang and Kam~Pui Chow.
\newblock Noise based deepfake detection via multi-head relative-interaction.
\newblock In {\em Proc. of the 37th AAAI}, volume~37, pages 14548--14556, 2023.

\bibitem{wang2021hififace}
Yuhan Wang, Xu~Chen, Junwei Zhu, Wenqing Chu, Ying Tai, Chengjie Wang, Jilin Li, Yongjian Wu, Feiyue Huang, and Rongrong Ji.
\newblock Hififace: 3d shape and semantic prior guided high fidelity face swapping.
\newblock {\em arXiv preprint arXiv:2106.09965}, 2021.

\bibitem{westerlund2019emergence}
Mika Westerlund.
\newblock The emergence of deepfake technology: A review.
\newblock {\em Technology innovation management review}, 9(11), 2019.

\bibitem{williams2009transmission}
David~B. Williams and C.~Barry Carter.
\newblock {\em Transmission Electron Microscopy}.
\newblock Springer, 2009.
\newblock See p.108.

\bibitem{news3}
Jessie Wu.
\newblock Face-swapping fraud sparks ai-powered crime fears in china.
\newblock Online, 2023-05-24.
\newblock \url{https://technode.com/2023/05/24/face-swapping-fraud-sparks-ai-powered-crime-fears-in-china/}.

\bibitem{xu2022MobileFaceSwap}
Zhiliang Xu, Zhibin Hong, Changxing Ding, Zhen Zhu, Junyu Han, Jingtuo Liu, and Errui Ding.
\newblock Mobilefaceswap: A lightweight framework for video face swapping.
\newblock In {\em Proc. of the 36th AAAI}, 2022.

\bibitem{yang2023avoid}
Wenyuan Yang, Xiaoyu Zhou, Zhikai Chen, Bofei Guo, Zhongjie Ba, Zhihua Xia, Xiaochun Cao, and Kui Ren.
\newblock Avoid-df: Audio-visual joint learning for detecting deepfake.
\newblock {\em IEEE Trans. on Information Forensics and Security}, 18:2015--2029, 2023.

\bibitem{yang2019exposing}
Xin Yang, Yuezun Li, and Siwei Lyu.
\newblock Exposing deep fakes using inconsistent head poses.
\newblock In {\em Proc. of the 44th IEEE ICASSP}, pages 8261--8265. IEEE, 2019.

\bibitem{yu2019attributing}
Ning Yu, Larry~S Davis, and Mario Fritz.
\newblock Attributing fake images to gans: Learning and analyzing gan fingerprints.
\newblock In {\em Proc. of the 15th ECCV}, pages 7556--7566, 2019.

\bibitem{yu2021survey}
Peipeng Yu, Zhihua Xia, Jianwei Fei, and Yujiang Lu.
\newblock A survey on deepfake video detection.
\newblock {\em Iet Biometrics}, 10(6):607--624, 2021.

\bibitem{zhang2022deepfake}
Tao Zhang.
\newblock Deepfake generation and detection, a survey.
\newblock {\em Multimedia Tools and Applications}, 81(5):6259--6276, 2022.

\bibitem{zhou2017two}
Peng Zhou, Xintong Han, Vlad~I Morariu, and Larry~S Davis.
\newblock Two-stream neural networks for tampered face detection.
\newblock In {\em Proc. of the 30th IEEE/CVF CVPRW}, pages 1831--1839. IEEE, 2017.

\bibitem{zhou2021face}
Tianfei Zhou, Wenguan Wang, Zhiyuan Liang, and Jianbing Shen.
\newblock Face forensics in the wild.
\newblock In {\em Proc. of the 34th IEEE/CVF CVPR}, pages 5778--5788, 2021.

\bibitem{zhou2021joint}
Yipin Zhou and Ser-Nam Lim.
\newblock Joint audio-visual deepfake detection.
\newblock In {\em Proc. of the 18th CVF/IEEE ICCV}, pages 14800--14809, 2021.

\end{thebibliography}
\end{document}